\documentclass[10pt,journal,compsoc]{IEEEtran}

\ifCLASSOPTIONcompsoc
\usepackage[utf8]{inputenc} 
\usepackage[T1]{fontenc}    
\usepackage{hyperref}
\usepackage{url}  
\usepackage{booktabs}
\usepackage{amsfonts}  
\usepackage{nicefrac} 
\usepackage{microtype}   
\usepackage{fancyhdr}      
\usepackage{bm}
\usepackage{amsmath}
\usepackage{algorithmic}
\usepackage{algorithm}
\usepackage{multirow}
\usepackage{graphicx}
\usepackage{subfigure}
\usepackage{wrapfig}
\usepackage{amsthm}
\usepackage{multicol}
\usepackage{amssymb}
\usepackage{pdfpages}
\usepackage{makecell}
\usepackage{xspace}
\usepackage{float}
\usepackage{threeparttable}
\usepackage{soul}
\usepackage{array}
\usepackage{cite}

\makeatletter
\DeclareRobustCommand\onedot{\futurelet\@let@token\@onedot}
\def\@onedot{\ifx\@let@token.\else.\null\fi\xspace}

\def\eg{\emph{e.g}\onedot} 
\def\ie{\emph{i.e}\onedot} 
 
\def\etc{\emph{etc}\onedot} 
 
\def\etal{\emph{et al}\onedot}
\makeatother

\allowdisplaybreaks[4]

\pdfoutput=1

\ifCLASSINFOpdf
\else
\fi

\begin{document}

\title{Contrastive Positive Sample Propagation \\along the Audio-Visual Event Line}

\author{Jinxing Zhou,
        Dan Guo*,
        and~Meng Wang*,~\IEEEmembership{Fellow,~IEEE}
\IEEEcompsocitemizethanks{
\IEEEcompsocthanksitem
J. Zhou, D. Guo, and M. Wang are with Key Laboratory of Knowledge Engineering with Big Data (HFUT), Ministry of Education, School of Computer Science and Information Engineering (School of Artiﬁcial Intelligence), Hefei University of Technology (HFUT), and Intelligent Interconnected Systems Laboratory of Anhui Province (HFUT), Hefei, 230601, China (e-mail: zhoujxhfut@gmail.com; guodan@hfut.edu.cn; eric.mengwang@gmail.com).
\IEEEcompsocthanksitem{
D. Guo and M. Wang are also with the Institute of Artificial Intelligence, Hefei Comprehensive National Science Center, Hefei, 230601, China.
}

\IEEEcompsocthanksitem{
Corresponding authors*: Dan Guo, Meng Wang.}
\IEEEcompsocthanksitem{
Code and dataset are available at \href{https://github.com/jasongief/CPSP}{https://github.com/jasongief/CPSP}.}
}
}

\markboth{IEEE transactions on pattern analysis and machine intelligence}
{Shell \MakeLowercase{\textit{et al.}}: Bare Advanced Demo of IEEEtran.cls for IEEE Computer Society Journals}

\IEEEtitleabstractindextext{
\begin{abstract}
Visual and audio signals often coexist in natural environments, forming audio-visual events (AVEs). Given a video, we aim to localize video segments containing an AVE and identify its category. It is pivotal to learn the discriminative features for each video segment. Unlike existing work focusing on audio-visual feature fusion, in this paper, we propose a new contrastive positive sample propagation (CPSP) method for better deep feature representation learning. The contribution of CPSP is to introduce the available full or weak label as a prior that constructs the exact positive-negative samples for contrastive learning.
Specifically, the CPSP involves comprehensive contrastive constraints:
pair-level positive sample propagation (PSP), segment-level and video-level positive sample activation (PSA$_S$ and PSA$_V$).
Three new contrastive objectives are proposed (\ie, $\mathcal{L}_{\text{avpsp}}$, $\mathcal{L}_ \text{spsa}$, and $\mathcal{L}_\text{vpsa}$) and introduced into both the fully and weakly supervised AVE localization. 
To draw a complete picture of the contrastive learning in AVE localization, we also study the self-supervised positive sample propagation (SSPSP).
As a result, CPSP is more helpful to obtain the refined audio-visual features that are distinguishable from the negatives, thus benefiting the classifier prediction. 
Extensive experiments on the AVE and the newly collected VGGSound-AVEL100k datasets verify the effectiveness and generalization ability of our method.
\end{abstract}

}

\maketitle

\IEEEdisplaynontitleabstractindextext
\IEEEpeerreviewmaketitle

\section{Introduction}\label{introduction}
\begin{figure}[t]
   \begin{center}
   \includegraphics[width=0.48\textwidth]{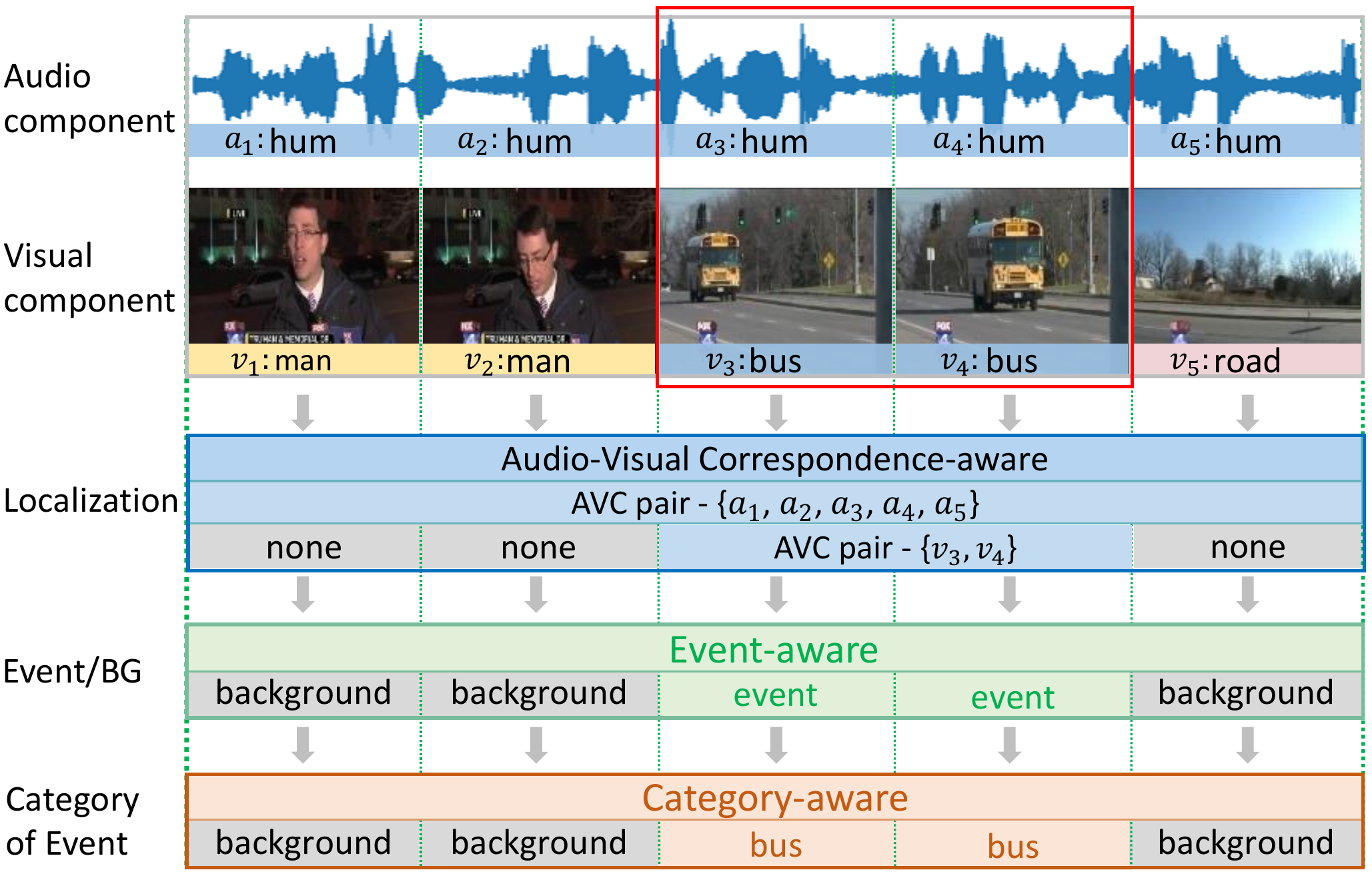}
   \end{center}
   \vspace{-0.4cm}
      \caption{An illustration of the AVE localization task.
      Each video segment is composed of an audio and a visual component.
      In this example,
      the ``hum'' of the bus exists in all the segments (audio modality), but
      the visual images of the ``bus'' appear only in the third and fourth segments (visual modality).
      So only these two segments (red boxes) are localized as \emph{bus event}, the remaining are recognized as \emph{background}.
      A localization system is expected to analyze and utilize the audio-visual pairs (audio-visual correspondence-aware), determine whether a video segment contains an audio-visual \emph{event} (event-aware), and further identify its event category (category-aware).
      }
   \label{fig:AVEL_task}
 \vspace{-3mm}
\end{figure}

\IEEEPARstart{A}{n} audio-visual event (AVE) often refers as an event that is both audible and visible in a video segment, \ie, a sound source appears in an image ({\em visible}) while the sound it makes also exists in the audio portion ({\em audible}). The AVE localization task is to find these video segments which contain an audio-visual event and classify it into a certain category.
We illustrate this task in Fig.~\ref{fig:AVEL_task}.
It belongs to the research topic of audio-visual scene understanding. It uses both audio and vision inputs to 
answer {\em if an event happens in both modalities at different video segments}.
The task must explore unconstrained videos (events in real life) that are not limited to the temporal consistency of lip reading or other human-making sounds. 

Recent literature has shown that by fusing multi-modality information can lead to better deep feature representation, \ie, audio-visual fusion~\cite{arandjelovic2017look} and text-visual fusion~\cite{radford2021learning}. However, building a large scale multi-modality pre-training datasets would require heavy manual labors to clean and annotate the raw video sets. To relief the manual labor, recent work either focuses on learning from noise supervision~\cite{cheng2019noise, jia2021scaling} or tries to automatically filter out irrelevant samples~\cite{tian2018audio}.
In this work, we devote to explore better deep feature representation for AVEs, which is served for the latter purpose. 

We study how to effectively leverage audio and visual information for event localization. In the current AVE localization work, two relations are considered: intra-modal and cross-modal relations. The former often addresses temporal relations in one single modality while the later also takes audio and visual relations into account. Pioneer works~\cite{lin2019dual, tian2018audio} often try to regress the class by directly concatenating features from synchronized audio-visual pairs;
their accuracy is often unsatisfying. The following works~\cite{wu2019dual, xu2020MM, xuan2020cross} utilize a self-attention mechanism to explicitly encode the temporal relations within intra-modality and some of them~\cite{ramaswamy2020makes, ramaswamy2020see, xu2020MM, xuan2020cross} also aggregate better audio-visual feature representations by encoding cross-modal relations. However, these methods aggregate all the audio and visual components, 
often ignore the interference caused by irrelevant audio-visual segment pairs during the fusion process. In this paper, we solve this problem in the cross-modal interaction by considering the relations from different perspectives: intra- and inter-videos. More importantly, we devote to feature aggregation and enhancement from high-relevant (positive) samples and can obtain better AVE localization accuracy.

About our observations, we argue and detail that the AVE localization task has three main challenges below.
(1) {\em Unconstrained audio-visual relevance matching}. On one hand, the sound-maker is often occluded by some event-irrelevant objects or even be out of the screen, \eg, the humming sound but accompanied by an announcer as shown in Fig.~\ref{fig:AVEL_task}. On the other hand, there are usually multiple objects contained in the visual scene which could be sound-makers or not, \eg, the humming accompanied by the bus, man, and road, also as shown in Fig.~\ref{fig:AVEL_task}; the audio signal is also inevitably mixed with other noises.
Such scenarios in unconstrained videos make it hard to match the audio-visual segment pairs in a flexible and accurate manner.
2) {\em Temporal inconsistency in AVE videos}. In real-life videos, the audio and visual signals are processed by an independent workflow.
This spawns the research on the audio-visual synchronization problem~\cite{chung2016out, korbar2018cooperative, Khosravan2019OnAM} and brings the temporal inconsistency issue. Such issue lets the segment event judgment (\ie, AVE or background) difficult. Especially for the weakly-supervised setting (refer Sec.~\ref{sec:problem_define} for the setting details), AVE localization task still asks to parse from segment-level but only given video-level label.
(3) {\em Distinction of similar but different representations.} For the AVE localization task, it not only requires locating the event along the timeline but also must identify its category. In order to obtain discriminative feature representations, we must constraint the representations learning to be category-aware, such as distinguishing the videos displaying musical instruments guitar and violin although these two events are with a negligible difference.

\begin{figure}[t]
   \begin{center}
     \includegraphics[width=0.48\textwidth]{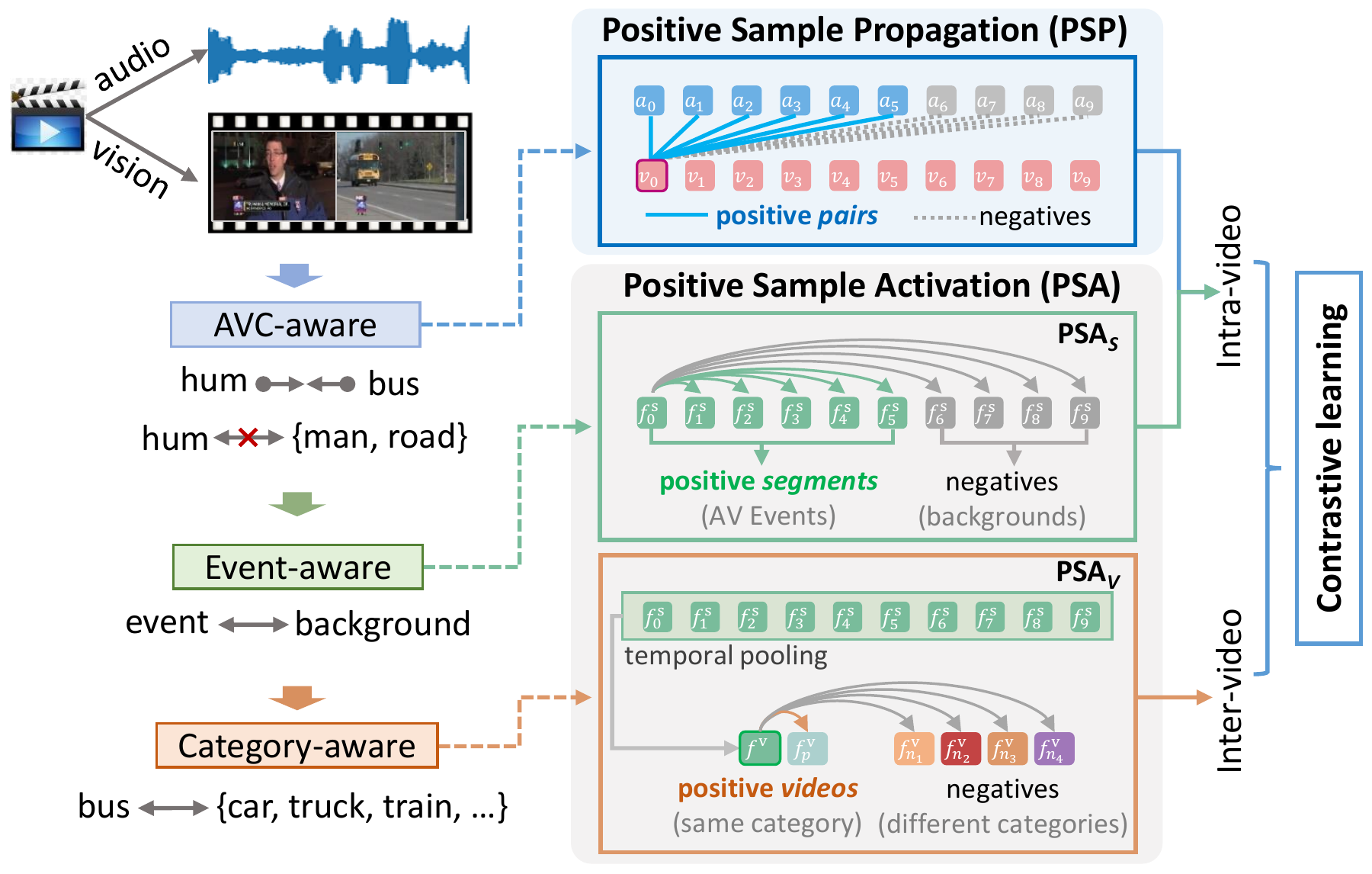}
   \end{center}
      \vspace{-0.6cm}
      \caption{An illustration of the proposed contrastive positive sample propagation (CPSP) method. The rounded rectangle in gray color represents a background segment, otherwise it means that the segment describes an event.
      The CPSP considers the AVE localization task from three aspects and works in a contrastive manner: 1) The \emph{pair-level} positive sample propagation (PSP) aims to select the most relevant audio-visual segment pairs for cross-modal feature aggregation (\emph{AVC-aware}). 2) The \emph{segment-level} positive sample activation (PSA$_S$) aims to distinguish the positive segments that contain an audio-visual event from the background segments (\emph{event-aware}). 3) The \emph{video-level} positive sample activation (PSA$_V$) activates the videos sharing the same event category as the positive samples (\emph{category-aware}).
      }
   \label{fig:HLPSA}
  \vspace{-0.3cm}
\end{figure}

\textbf{The proposed Contrastive Positive Sample Propagation model (CPSP).} To deal with aforementioned challenges, as shown in Fig.~\ref{fig:HLPSA}, we propose a new {\bf \em CPSP method}
that enables the localization system to encode discriminative representations by activating the \emph{positive instances in the audio-visual data} from three levels, \ie, the most relevant audio-visual pairs (\emph{pair-level}), the segments containing an AVE consistently in audio and vision modalities (rather than the background, \emph{segment-level}), the videos belonging to the same event category (rather than other categories, \emph{video-level}). By exploiting these positive samples, the learned audio-visual representation encourages our model to be AVC (audio-visual correspondence)-aware, event-aware, and category-aware, which exactly matches the goal of AVE localization task. We introduce the details next.

Specifically, as shown in Fig.~\ref{fig:system_flow}, we propose a new {\bf \em Positive Sample Propagation (PSP) module}. In a nutshell, PSP constructs an all-pair similarity map between each audio and visual segment and cuts off the entries that are below a pre-set similarity threshold, and then aggregates the audio and visual features without considering the negative and weak entries in an online fashion. Through various visualizations, we show that the PSP allows the most relevant features that are not necessarily synchronized to be aggregated in an online fashion.
It is noteworthy that our PSP is \emph{AVC-aware}, which is different from existing cross-modal feature fusion methods~\cite{ramaswamy2020makes, ramaswamy2020see, xu2020MM, xuan2020cross}.
A concise illustration of the PSP has been shown in Fig.~\ref{fig:HLPSA} and details can be seen from Fig.~\ref{fig:PSP_process}.

Apart from the PSP, it is also significant to effectively address the difficulties: 1) complex visual or audio backgrounds in an unconstrained video make it difficult to localize an AVE along the time line, and 2) localizing and recognizing an AVE category requires the model to deeply exploit the representative features among different AVE videos. Thus, as illustrated in Fig.~\ref{fig:system_flow}, we propose a new {\bf \em Positive Sample Activation (PSA) module} to constraint the representation of target segments containing an AVE to be possibly distinguishable from both the backgrounds in the same video (intra-video) and the segments of other videos with different event categories (inter-video). Specifically, the PSA is conducted from the segment-level (PSA$_S$) and video-level (PSA$_V$) whereas the PSA$_S$ is designed to address the former purpose, while the PSA$_V$ is proposed to solve the latter. As illustrated in Fig.~\ref{fig:HLPSA}, we show that the PSA$_S$ is \emph{event-aware} and the PSA$_V$ is \emph{category-aware}. Details are introduced in Sec.~\ref{sec:contrastive_learning}.

It is nontrivial to build positive and negative connections between complex visual scenes and intricate sounds. We utilize the principle of the {\em contrastive learning} without additional annotations to optimize the audio-visual representation learning. Three new contrastive losses, \ie, the $\mathcal{L}_{\text{avpsp}}$ for the PSP, $\mathcal{L}_{\text{spsa}}$ for the PSA$_S$, and $\mathcal{L}_{\text{vpsa}}$ for the PSA$_V$, are proposed for gathering positive samples while pushing the negatives away from them in the feature space.
They are different from these audio-visual related works~\cite{afouras20ssl, ma2021contrastive, wu2021exploring} that take the synchronized audio-visual segments as positive samples, which is not compatible with AVE localization. In our work, we explore the positive (high relevant) AVC pairs and consider performing contrastive techniques from more levels, \ie, adding the segment-level and video-level contrasting.
To the best of our knowledge, we are the first to introduce contrastive learning to solve the AVE localization problem and provide a comprehensive discussion about the three levels of positive instances covering both intra- and inter- AVE video correlations as shown in fig.~\ref{fig:HLPSA}.

\textbf{The improvement of backbone for fully and weakly supervised AVE localization.} Besides the proposed CPSP,
we analyze the fully and weakly supervised AVE localization, and further propose two improvements that work under each setting, respectively.
On the one hand, an audio-visual pair similarity loss based on the PSP $\mathcal{L}_{\text{avpsp}}$ is introduced under the fully supervised setting that encourages the network to learn high correlated features for the audio-visual pair if they contain the same event. 
On the other hand, we propose a weighting branch in the weakly supervised setting, which gives temporal weights to the segment features. We evaluate these two improvements on the standard AVE dataset~\cite{tian2018audio} and the experimental results demonstrate their effectiveness. It is noteworthy that these two improvements are not only helpful in the proposed CPSP method but also can be applied to other localization networks (relevant results have been shown in Sec.~\ref{sec:evaluation_improvements}).

To summarize, the proposed CPSP including the PSP and PSA modules devotes to better deep representation by imposing contrastive constraints on the localization system from three semantic levels (positive audio-visual instances of AVC pairs, segments, and videos), which covers both intra- and inter- AVE video correlations. The main contributions are summarized as follows:
\begin{itemize}
\item The proposed PSP allows to explore the most relevant (positive) \emph{pair-level} features that are not necessarily synchronized but semantic-related to be aggregated and enables to encode more distinguishable audio-visual representations.
\item The proposed PSA explicitly activates the positive samples from additional \emph{segment-level} and \emph{video-level} rather than directly sending the fused audio-visual features into the final classifier such as in previous works\cite{xu2020MM, xuan2020cross, ramaswamy2020makes, ramaswamy2020see}.
\item The improvements of backbone proposed for the fully and weakly supervised settings consistently benefit our localization system and are also advantageous in other networks.
\item Extensive experiments demonstrate the effectiveness of following designs and our method achieves new state-of-the-art performances under both settings.
\end{itemize}

At last, we remind that the PSP module was first introduced in our previous work~\cite{zhou2021positive}.
Compared to the preliminary version, in this paper, we have made improvements in six aspects:
(1) in addition to the PSP, we expand it to the CPSP by adding the Positive Sample Activation (PSA, including PSA$_S$ and PSA$_V$) scheme that systematically exploiting segment-level and video-level positive audio-visual instances; (2) we perform a comprehensive survey of relevant works about the contrastive learning in the field of audio-visual representation learning in Sec.~\ref{sec:related_work}; (3) two new contrastive objective losses are designed and introduced into the AVE localization in Sec.~\ref{sec:contrastive_learning}; we add the discussion about the objectives in Sec.~\ref{sec:discussion};
(4) we implement a self-supervised contrastive method SSPSP and give more analysis by comparing it with the CPSP in Sec.~\ref{sec:quanti2}.
(5) we release a large-scale VGGSound-AVEL100k dataset for AVEL task.
The videos are sampled from VGGSound~\cite{chen2020vggsound} where the video-level categories are given and we provide segment-level annotations.
Considerable experiments are conducted on this large dataset to evaluate our model and more detailed analyses are provided in Sec.~\ref{sec:quanti2}.
(6) we extend the proposed method on the LLP~\cite{tian2020avvp} dataset collected for a similar but more challenging audio-visual video parsing (AVVP) task in Sec.~\ref{sec:avvp} and the results also demonstrate the effectiveness and generalization ability of our method.
In brief, the CPSP proposed in this paper makes the localization framework much more comprehensive and extensive experiments make the CPSP more convincing.

The rest of this paper is organized as follows. Sec.~\ref{sec:related_work} provides an overview of the related works. Sec.~\ref{sec:problem_define} introduces two settings of the AVE localization problem, \ie, the fully and weakly supervised tasks. Sec.~\ref{sec:method} elaborates on the proposed CPSP. Discussions on the CPSP methodology are shown in Sec.~\ref{sec:discussion}. The experimental results and analyses are presented in Sec.~\ref{sec:experiment}, and conclusion are given in Sec.~\ref{sec:conclusion}.

\vspace{-3mm}
\section{Related Work}\label{sec:related_work}

\textbf{Audio-visual correspondence (AVC)} aims to predict whether a given visual image corresponds to a short audio recording.
The task is asked to judge whether the audio and visual signals describe the same object,
\eg, dog {\em v.s.} bark, cat {\em v.s.} meow.
It is a self-supervised problem since the visual image is usually accompanied by the corresponding sound in video data.
Existing methods try to evaluate the correspondences by measuring the audio-visual similarity~\cite{arandjelovic2017look, arandjelovic2018objects, aytar2016soundnet, cheng2020look, fayek2020large}.
It will get a large similarity score if the audio-visual pair is corresponding, otherwise, a low score.
This is in line with our focused AVE localization problem since the synchronized audio-visual pair of a target {\em event} segment must be corresponding.
The difference is that AVC tackles the correspondence of an audio and an image, rather than an audio and a video in our AVE localization task.
So, we are motivated to tackle the abundant audio-visual segment pairs in AVE localization problem by further exploring and exploiting the audio-visual similarity.

\textbf{Sound source localization (SSL)} aims to localize those visual regions which are relevant to the provided audio signal.
It is related to {\em sound source separation} problem, which mainly focuses on the event of people speech~\cite{darrell2000audio, xu2018single, afouras2018deep, jenrungrot2020cone, afouras20ssl, gao2021visualvoice} or musical instrument playing~\cite{parekh2017motion, pu2017audio, zhao2018sound, gao2018learning, zhao2019sound, gao2019co}.
For SSL, there is usually a condition that the sound source must appear in the visual image.
In other words, it mainly focuses on the \emph{visual} localization while the AVE localization devotes to the \emph{temporal} localization.
SSL has two settings: the single and multiple sound source(s) localization.
For the single SSL, the localization map can be easier obtained in an unsupervised manner by directly computing the similarity between the audio feature and visual feature map.
The multiple SSL is more challenging that requires to accurately locate the sound-maker when there are multiple sound sources~\cite{qian2020multiple, hu2019deep, hu2020discriminative, zhou2022avs}.
The class activation mapping (CAM)~\cite{zhou2016CAM} is helped to realize class-aware object localization.
Qian \etal~\cite{qian2020multiple}
adapt the Grad-CAM~\cite{selvaraju2017grad} to disentangle class-specific features for multiple SSL.
Hu \etal~\cite{hu2019deep} maps audio and visual features into respective K cluster centers and take the center distance as a supervision to rank the paired audio-visual objects.
Hu \etal~\cite{hu2020discriminative} first learn the object semantics in single SSL then use that to help with multiple SSL.
Recent methods use contrastive techniques to utilize the discriminative sound characteristics and diverse object appearances.
Senocak \etal~\cite{senocak2021TAPMI} propose a triplet loss working in an unsupervised manner.
Afouras \etal~\cite{afouras20ssl} utilize
a contrastive loss to train the model in a self-supervised learning way.
Both of these methods~\cite{afouras20ssl, senocak2021TAPMI}
need to construct positive and negative audio-visual pair samples.
Considering the positive and negative samples can also be obtained in AVE localization task,
we propose to exploit the abundant audio-visual instances for contrastive learning.

\textbf{Audio-visual event localization (AVEL)} aims to distinguish those segments including an
audio-visual event from a long video.
Different from the acoustic event classification~\cite{hershey2017cnn, kong2018audio, kumar2018knowledge, mcfee2018adaptive} or video classification~\cite{karpathy2014large, long2018attention, long2018multimodal, wang2018appearance, tran2019video} making a prediction based on the whole audio or video embedding, AVE localization requires to judge the audio-visual correspondence and event category for each segment.
Currently, the AVEL task is a supervised problem with weak labels or full labels. The former merely contains video-level labels, and the latter refers to both segment-level and video-level annotations. 
Existing works mainly focus on the audio-visual fusion process.
A dual multimodal residual network is proposed in~\cite{tian2018audio}.
Lin \etal~\cite{lin2019dual} adapt a bi-directional LSTM~\cite{schuster1997bidirectional} to fuse audio and visual features in a seq2seq manner.
These methods simply concatenate the synchronized audio-visual features during fusion.
Subsequent work utilizes a bilinear method~\cite{ramaswamy2020makes, ramaswamy2020see} or a joint co-attention strategy~\cite{xuan2020cross, Duan_2021_WACV} to capture cross-modal relations
between both synchronized and unsynchronized audio-visual pairs.
Self-attention mechanism is also widely used to encode temporal relation in both the AVEL~\cite{wu2019dual, xu2020MM, xuan2020cross} and \emph{audio-visual video parsing (a new weakly supervised audio-visual related task)}~\cite{tian2020avvp, wu2021explore, lin2021exploring}.
Lin \etal~\cite{Lin_2020_ACCV} design an audio-visual transformer to describe local spatial and temporal information.
The visual frame is divided into patches and adjacent frames are utilized, making the model complicated and computationally intensive.
Xu \etal~\cite{xu2020MM} attempt to use the concatenating audio-visual features as the supervision
then the feature of each modality is updated by separate modules.
Unlike these, the proposed CPSP method has a further in-depth study on the abundant audio-visual pairs, event and background segments, and similar videos but with different categories, activating the most relevant ones.
Relying on these positive samples, more distinguished audio-visual features can be obtained after feature aggregation.

\textbf{Contrastive learning in audio-visual field.}
The technique of contrastive learning turns out to be an effective solution that is widely used in self-supervised learning~\cite{oord2018CPC, he2020momentum, chen2020simple, morgado2021audio, harwath2018jointly} and various weakly supervised tasks~\cite{zhang2020counterfactual, zhang2021cola, gupta2020contrastive, ki2020sample}. 
Seeing data in a large batch size, the model learns discriminative representations by identifying the positive or negative samples. The key of contrastive learning is how to construct the positive and negative samples. 
Recently, some researchers start to explore injecting the label information to accurately select more positive/negative samples for better representation learning. 
Such supervised contrastive learning has shown its superiority in both computer vision~\cite{SCL2020, zeng2021modeling} and natural language processing tasks~\cite{sedghamiz2021supcl, gunel2020supervised}.
In the audio-visual related works~\cite{afouras20ssl, ma2021contrastive, wu2021exploring}, they usually adopt the self-supervised contrastive manner, \ie, directly selecting the positive sample from the synchronized audio-visual segment and contrast them with the negatives come from different timestamps.
However, this is not compatible with AVE localization: an audio and visual segment can be regarded as a positive sample as long as they describe the \emph{same event}, and vice versa.
To effectively distinguish an AVE, it is vital to construct exact positive and negative samples from the video segments for contrastive learning.
Inspired by these, we propose the PSA in a supervised manner that explicitly performs contrastive learning between segments and videos with the segment/video-level label prior.
To the best of our knowledge, we are the first to utilize the contrastive learning to solve AVE localization problem and we explore comprehensive contrastive strategies from different levels.

\section{Problem statement}\label{sec:problem_define}
AVE localization aims to find out those segments
containing an audio-visual event \cite{tian2018audio}.
In other words, AVE localization is expected to decide whether each synchronized audio-visual pair depicts an event.
Besides, AVE localization needs to identify the event category for each segment.
Specifically, a video sequence $S$ is divided into $T$ non-overlapping yet continuous segments $\{S_t^v, S_t^a\}^{T}_{t=1}$, and each segment is one-second in length. $S^v$ and $S^a$ are the visual and audio components, respectively. We consider two settings of this task, to be described below.

\textbf{Fully-supervised AVE localization.}
Under the fully supervised setting, the event label of every video segment is given, indicating
whether the segment denotes an event and which category the event belongs to.
We denote the event label of the $t^{\text{th}}$ segment as $\bm{y}_t = { \{ y_t^c | y_t^c \in \{0, 1\},  \sum_{c=1}^{C} y_t^c = 1\} \in \mathbb{R}^C}$,
where $C$ is the number of categories (including the {\em background}).
Then, the label for the entire video can be written as $\bm{Y}^{\text{fully}} = [\bm{y}_1; \bm{y}_2; ...; \bm{y}_T] \in \mathbb{R}^{T \times C}$.
Through $\bm{Y}^{\text{fully}}$, we know whether an arbitrary synchronized audio-visual pair at time $t$ is an event: if the $1$ of its event label $\bm{y}_t$ is at the entry of a certain event instead of the \emph{background},
the pair describes an event and otherwise does not.

\begin{figure*}[t]
  \begin{center}
      \setlength{\abovecaptionskip}{0.cm}
    \includegraphics[width=\textwidth]{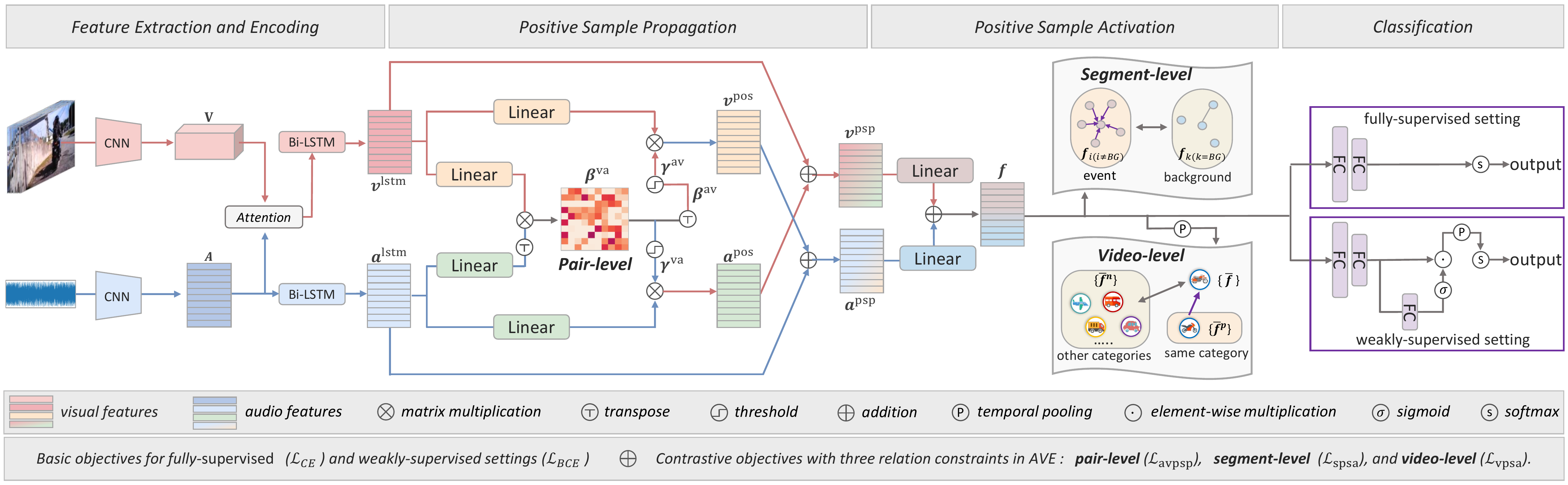}
  \end{center}
  \vspace{-0.5cm}
      \caption{System Flow. We first extract and encode video and audio features through existing modules such as AVGA~\cite{tian2018audio} and Bi-LSTM.
      The proposed {\em positive sample propagation} (PSP) takes the LSTM encoded features as input;
      an affinity matrix is computed before selecting the connections of audio-visual segment pairs using thresholding.
      In this module, 
      audio and visual features are aggregated by feature propagation through the {\em pair-level positive} connections.
      Then, the fused features are further processed under the constraints of the proposed {\em positive sample activation} (PSA), where the {\em segment-level} (PSA$_S$) enforces features of these video segments containing the same event ({\em positive}) to be possibly closer and be away from the background segments while the {\em video-level} (PSA$_V$) encourages representations of the videos sharing the same event category ({\em positive}) to be similar. Details are described in Sec.~\ref{sec:contrastive_learning}.
      In the last stage, we classify the event into predefined categories.
      For the supervised setting, apart from the commonly used cross-entropy (CE) loss, we further propose an audio-visual pair similarity loss based on the PSP which enforces similar features between them when they contain an event. For the weakly supervised setting, we introduce another FC layer that gives weights to different video segments: higher weights are given event-containing segments.
      The whole system is optimized by the basic classification loss with additional proposed contrastive objectives, detailed in Sec.~\ref{sec:objective_function}.
      }
      \vspace{-0.3cm}
  \label{fig:system_flow}
\end{figure*}

\textbf{Weakly-supervised AVE localization.}
We adapt the weakly-supervised setting following \cite{lin2019dual,xuan2020cross}, where the label $\bm{Y}^{\text{weak}} \in \mathbb{R}^{1 \times C}$ is the average pooling value of $\bm{Y}^{\text{fully}}$ along the column.
It implies the proportion of video segments that contain an event. 
This setting is different from the fully supervised one because the event label of each segment $y_t$ is unknown, making the problem more challenging.

\section{Our method}\label{sec:method}

The overall pipeline of our system is illustrated in Fig.~\ref{fig:system_flow},
which includes four modules:
a feature extraction and encoding module (Sec.~\ref{sec:feature_extractor}),
a positive sample propagation module (Sec.~\ref{sec:PSP}), a
positive sample activation module (Sec.~\ref{sec:contrastive_learning}), and a classification module (Sec.~\ref{sec:classification}).
In the {\em feature extraction and encoding} module,
the audio-guided visual attention (AVGA~\cite{tian2018audio}) is adapted for early fusion to make the model focus on those visual regions closely related to the audio component.
Then Bi-LSTM is utilized to encode temporal relations in video segments for each modality.
The LSTM encoded audio and visual features are sent to the proposed {\em positive sample propagation (PSP)} module.
PSP is able to select those positive connections of audio-visual segment pairs by measuring the cross-modal similarity with thresholding, $i.e.$, the pair-level contrastive constraint. 
Audio and visual features are aggregated by feature propagation through the positive connections.
The fused audio-visual features after PSP are further processed by two contrastive constraints, {\em i.e.}, the \emph{
positive sample activation} from both segment-level (PSA$_S$) and video-level (PSA$_V$), which refine the audio-visual features for segments containing an event in a video and different videos but sharing the same event category, respectively.
The updated features are then sent to the final {\em classification} module, predicting which video segments contain an event and the event category.

\subsection{Feature extraction and encoding}\label{sec:feature_extractor}

The visual and synchronized audio segments are processed by pretrained convolutional neural networks (CNNs).
We denote the resulting visual feature as $ \pmb V \in \mathbb{R}^{T \times N \times d_v } $,
where $d_v$ is the feature dimension, $N = H \times W$, $H$ and $W$ are the height and width of the feature map, respectively.
The extracted audio feature is denoted as $ \pmb A \in \mathbb{R}^{T \times d_a} $, where $d_a$ denotes feature dimension.
We then directly adapt AGVA~\cite{tian2018audio} for multi-modal early fusion. AVGA allows the model to focus on visual regions that are relevant to the audio component.
To encode the temporal relationship in video sequences, the visual and
audio features after AVGA are further sent to two independent Bi-LSTMs.
The updated visual and audio features are
represented as $\bm{v}^{\text{lstm}}\in \mathbb{R}^{T \times d_l}$ and $\bm{a}^{\text{lstm}} \in \mathbb{R}^{T \times d_l}$, respectively.

\subsection{Positive sample propagation (PSP)}\label{sec:PSP}
PSP allows the network to learn more representative features
by exploiting the similarities of synchronized and unsynchronized audio-visual segment pairs.
It involves three steps.

\begin{figure}[t]
   \begin{center}
   \setlength{\abovecaptionskip}{0.cm}
   \includegraphics[width=0.45\textwidth]{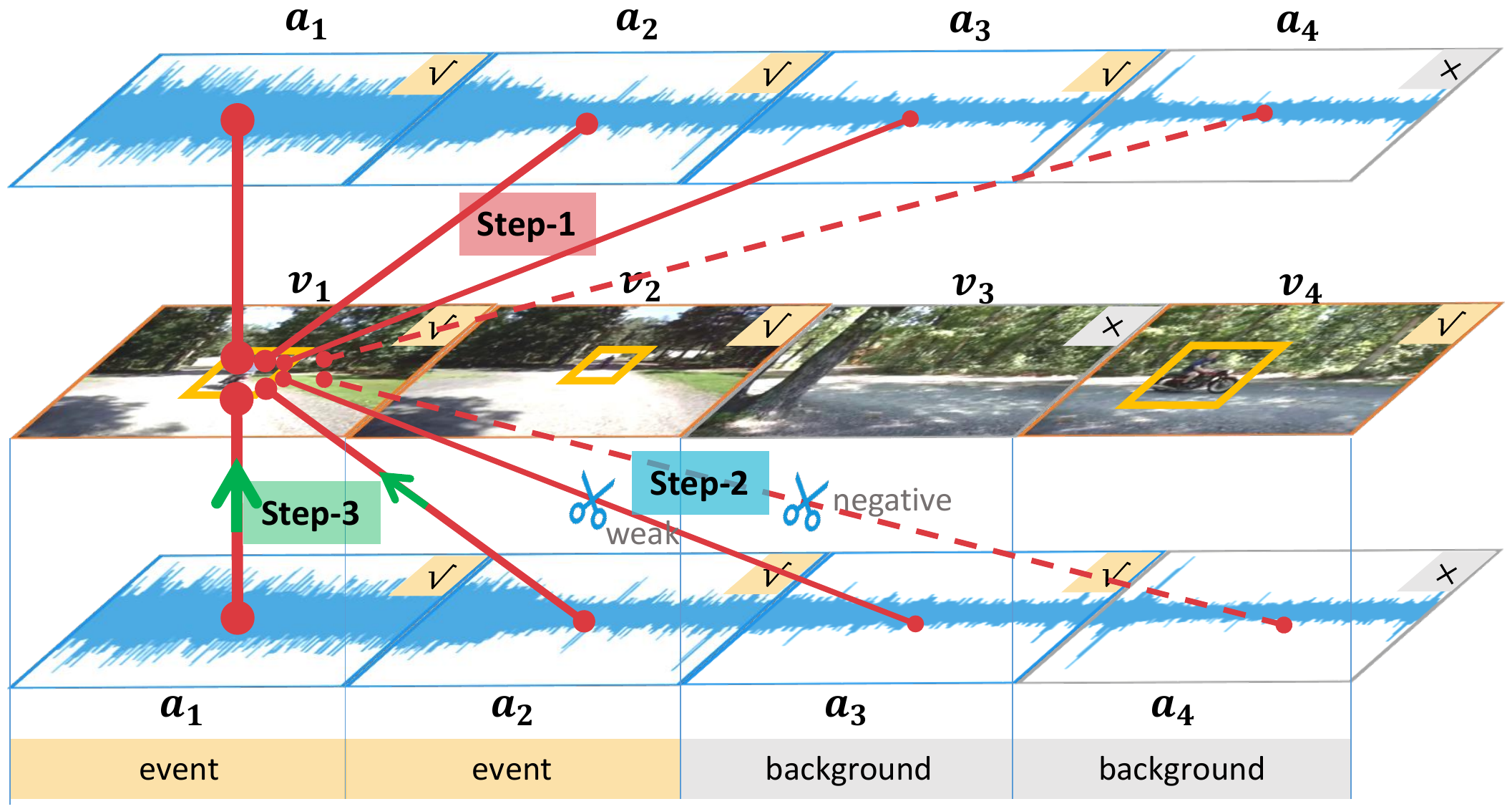}
   \end{center}
   \vspace{-0.5cm}
      \caption{An illustration of the proposed PSP.
      In this example, only the first two video segments contain an audio-visual event, \ie, \emph{motorcycle}.
      ``$\surd$'' denotes the audio or visual segment that describes the event, while ``$\times$'' means not.
      The red lines denote connections of audio-visual pairs,
      solid lines represent connections formed by relevant pairs,
      while dotted lines denote irrelevant pairs.
      The thickness of line reflects the similarity of the audio-visual pair.
      $v_1 \leftrightarrow a_4$ is a {\em negative} connection, formed by irrelevant audio-visual pair with negative similarity value.
      $v_1 \leftrightarrow a_3$ and $v_1 \leftrightarrow a_1/a_2$ are {\em weak} and {\em positive connections} respectively,
      determined via similarity.
      ``Step-1'' corresponds to the all-pair connection construction,
      while ``Step-2'' denotes the pruning of the negative and weak connections,
      and the green arrow indicates the {\em positive} direction of feature propagation in ``Step-3''.
      }
   \label{fig:PSP_process}
\vspace{-0.3cm}
\end{figure}

In \emph{all-pair connection construction},
all the audio-visual pairs are connected.
As shown in Fig.~\ref{fig:PSP_process},
here we only display the connections of one visual segment for simplicity,
{\em i.e.}, $\langle v_1 \leftrightarrow a_1/a_2/a_3/a_4 \rangle$.
The strength of these connections are measured
by the similarity between the audio-visual components $\langle{\bm{a}^{\text{lstm}}, \bm{v}^{\text{lstm}}} \rangle$,
computed by,
\begin{equation}
   \bm{\beta}^{\text{va}} = \frac{(\bm{v}^{\text{lstm}}{\bm{W}_1^v})(\bm{a}^{\text{lstm}}{\bm{W}_1^a})^{\top}} {\sqrt{d_l}}, \quad
   \bm{\beta}^{\text{av}} = (\bm{\beta}^{\text{va}})^{\top},
\end{equation}
where $\bm{W}_1^v \mbox{ and } \bm{W}_1^a \in \mathbb{R}^{d_l \times d_h}$
are learnable parameters of linear transformations,
implemented by a linear layer, and
$d_l$ is the dimension of the audio or visual feature. 
$\bm{\beta}^{\text{va}} \mbox{ and } \bm{\beta}^{\text{av}} \in \mathbb{R}^{T \times T}$ are the similarity matrices.

Second, we \emph{prune the negative and weak connections}.
Specifically, the connections constructed in the first step
are divided into three groups according to the similarity values: negative, weak, and positive.
As a classification task, the success of AVE localization highly depends on the richness and correctness of training samples for each class. That is, we aim to collect possibly many and relevant \emph{positive} connections. 
We achieve this goal by filtering out the weak and negative ones, \eg, $v_1 \leftrightarrow a_3$ and $v_1 \leftrightarrow a_4$ as shown in Fig.~\ref{fig:PSP_process}.
We begin with processing all the audio-visual pairs
with the ReLU activation function,
cutting off connections with negative similarity values.
Row-wise $\ell_1$ normalization is then performed,
yielding the normalized similarity matrices
$\bm{\beta}^{\text{va}}$ and $\bm{\beta}^{\text{va}}$.

\begin{figure*}[t]
\subfigure[Illustration of the {\em segment-level} positive sample activation (PSA$_S$). On the \textbf{left},
only the first five video segments contain the event \emph{aircraft};
for these segments, take anyone as an \emph{anchor}, the remaining four segments can be regarded as its \emph{positive} samples, while the last five segments constitute the \emph{negative} samples.
Next, as illustrated on the \textbf{right}, in PSA$_S$, the positive samples are gathered around the anchor while pushed away from the negative ones.]{
  \centering
  \includegraphics[width={\textwidth}]{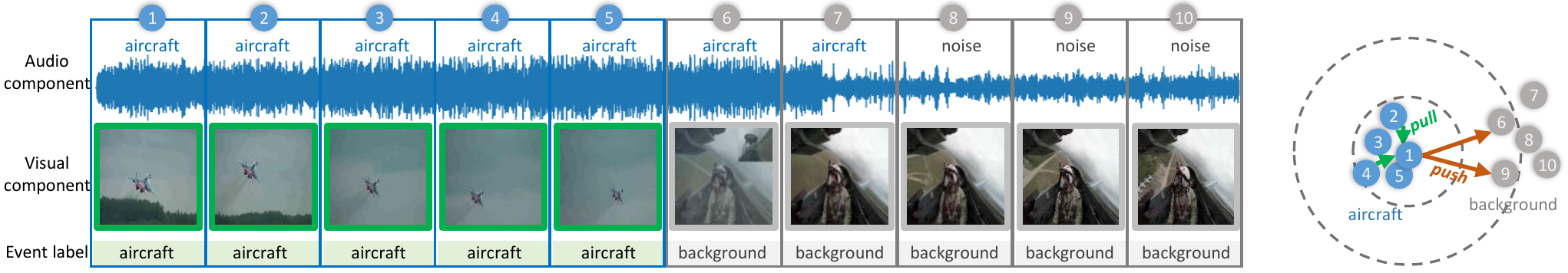}  
  \label{fig:subset_bg}
}
\vspace{-2mm}
\newline
\subfigure[Illustration of the {\em video-level} positive sample activation (PSA$_V$). On the \textbf{left},
all the segments in the video contain the event \emph{guitar}.
Videos sharing the same event category \emph{guitar} are treated as the \emph{positive} samples, while the \emph{negative} samples come from those holding other event categories.
As illustrated on the \textbf{right}, for an anchor video, we merely select one positive sample belonging to the same event category with the largest Euclidean distance, and negative samples classified to other categories but with top-$K$ smallest distances.
The positive and top-$K$ hard negative samples are online selected.
]{
  \centering
  \includegraphics[width={\textwidth}]{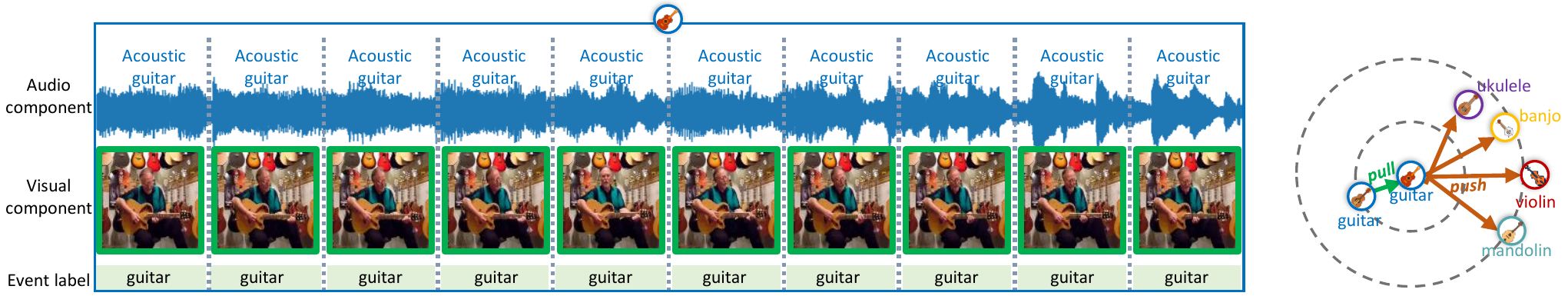}  
  \label{fig:subset_ae}
}
\vspace{-5mm}
\caption{We show two types of video data and illustrate the 
contrastive constraints (PSA$_S$ and PSA$_V$) that are proposed to further
exploit more positive samples from both segment-level and video-level, respectively. The green boxes represent an \emph{event} happens in the visual segment, while gray boxes mean not.}
\vspace{-2mm}
\label{fig:contrastive_strategy}
\end{figure*}

The negative and weak connections are presumably featured by smaller similarity values, so we simply adapt a thresholding method, written as, 
\begin{equation}
\begin{split}
   \bm{\gamma}^{\text{va}} = \bm{\beta}^{\text{va}} \mathbb{I}(\bm{\beta}^{\text{va}}-\tau), \\
   \bm{\gamma}^{\text{av}} = \bm{\beta}^{\text{av}} \mathbb{I}(\bm{\beta}^{\text{av}}-\tau),
   \label{threshold_operation_1}
\end{split}
\end{equation}
where $\tau$ is the hyper-parameter, controlling how many connections will be pruned. $\mathbb{I(\cdot)}$ is an indicator function, which outputs $1$ when the input
is greater than or equal to $0$, and otherwise outputs $0$. After thresholding, row-wise $\ell_1$ normalization is again performed to obtain the final similarity matrices $\bm{\gamma}^{\text{va}},$ $\bm{\gamma}^{\text{av}}\in \mathbb{R}^{T \times T}$.

\emph{Online feature aggregation}. The above step identifies audio (visual) components with high similarities with a given visual (audio) component, \eg, $v_1 \leftrightarrow a_1$ and $v_1 \leftrightarrow a_2$ shown in Fig.~\ref{fig:PSP_process}. This is essentially a positive sample propagation process that can be utilized to update the features of audio or visual components. 
Particularly, given the connection weights $\bm{\gamma}^{\text{av}}$ and $\bm{\gamma}^{\text{va}}$, the audio and visual features $\bm{a}^{\text{psp}}$ and $\bm{v}^{\text{psp}}$  are respectively updated as,
\begin{equation}
   \begin{split}
       \bm{a}^{\text{psp}} = \overbrace {\bm{\gamma}^{\text{av}}(\bm{v}^{\text{lstm}}{\bm{W}_2^v})}^{\bm{v}^{\text{pos}}} + \bm{a}^{\text{lstm}},\\
       \bm{v}^{\text{psp}} = \overbrace {\bm{\gamma}^{\text{va}}(\bm{a}^{\text{lstm}}{\bm{W}_2^a})}^{\bm{a}^{\text{pos}}} + \bm{v}^{\text{lstm}},
   \end{split}
   \label{eq:3}
\end{equation}
where $\bm{W}_2^{a}, \bm{W}_2^{v} \in \mathbb{R}^{d_l \times d_l}$ are parameters defining linear transformations, and
$\bm{a}^{\text{psp}}, \bm{v}^{\text{psp}}  \in \mathbb{R}^{T \times d_l}$.

Generally, the audio (visual) feature $\bm{a}^{\text{psp}}$ ($\bm{v}^{\text{psp}}$) is enhanced by the propagated positive support from the other modality. 
This practice allows us to learn more discriminative audio-visual representations, displayed in Fig.~\ref{fig:feature_distribution}. More discussions are provided in Sec. \ref{sec:discussion}.

\subsection{Positive sample activation (PSA)}\label{sec:contrastive_learning} 
PSA is designed to make the model more event-aware and category-aware. It involves two steps.
We activate more positive samples from both segment and video levels. We introduce the PSA$_S$ and PSA$_V$ with contrastive strategies below. Before that, we first fuse the audio and visual features $\bm{a}^{\text{psp}}$ and $\bm{v}^{\text{psp}}$ into an integrated audio-visual feature $\bm{f}$ as follows:
\begin{equation}\label{eq:av_fusion}
  \bm{f} = \frac {1}{2}[{\mathcal{N}(\bm{v}^{\text{psp}}{\bm{W}_3^v}) + \mathcal{N}({\bm{a}^{\text{psp}}}{\bm{W}_3^a})}],
\end{equation}
where $\bm{f}\in \mathbb{R}^{T \times d_l}$ is the feature of video segments, $ \mathcal{N}(\cdot) $ represents layer normalization,
$ \bm{W}_3^v, \bm{W}_3^a \in \mathbb{R}^{d_l \times d_l} $ represent learnable parameters in the linear layers. $\bm{f}$ can be used to represent the segment feature and be summarized to the video feature.

\textbf{\subsubsection{Segment-level positive sample activation (PSA$_S$).}}
To make the model be event-aware, we design a contrastive strategy from the segment-level.
As shown in Fig.~\ref{fig:subset_bg}, there are two sets of segments: segments depicting an audio-visual event constitute the \emph{event set}, while remaining segments form the \emph{background set}. 
We present a contrastive strategy to perceive the difference between these two video segment sets.
Take arbitrary segment from the event set as an anchor, the remaining ones in the event set are regarded as its \emph{positive} samples, while the segments in the background set are treated as \emph{negative} samples.
As shown in the right of Fig.~\ref{fig:subset_bg}, positive samples should be pulled together to the anchor, while the negative ones are pushed away. 
The segment-level contrastive objective takes the following form,
\begin{equation}\label{eq:loss_scon}
\begin{gathered}
\begin{split}
    &\mathcal{L}_{\text{spsa}} \!= \\
    & \!-\!
    \frac{1}{N_i^e}\! \sum_{i=1}^{N_i^e}\! \log(\frac{\text{exp}(\frac{\text{sim}(\bm{f}_i, \bm{f}_j)}{\eta})}{\text{exp}(\frac{\text{sim}(\bm{f}_i, \bm{f}_j)}{\eta})\! + \!\frac{1}{N_i^{\text{bg}}}\!\sum_{k=1}^{N_i^{\text{bg}}}\!\text{exp}(\frac{\text{sim}(\bm{f}_i, \bm{f}_k)}{\eta})}\!), \\
\end{split}
\end{gathered}
\end{equation}
where features $\bm{f}_i$ and $\bm{f}_j$ belongs to the event set ($i \ne j$), $\bm{f}_i$ is the segment anchor, $\bm{f}_j$ is one of $\bm{f}_i$'s \emph{positive} samples, $\bm{f}_k$ denotes a \emph{negative} sample comes from the background set.
$N_i^e$ and $N_i^{\text{bg}}$ are the total numbers of the event and background segments, respectively.
$\text{sim}(\cdot, \cdot)$ computes the dot product of the $\ell_2$ normalized vectors (\ie, cosine similarity);
$\eta$ is a temperature parameter controlling the concentration level of feature distribution.

\textbf{\subsubsection{Video-level positive sample activation (PSA$_V$).}}
To make the model be category-aware, we design another online contrastive strategy from video-level. As shown in Fig.~\ref{fig:subset_ae}, there are some instrument-related events in dataset ({\em e.g.} guitar, violin, mandolin, banjo, ukulele), which are hard to distinguish since they are similar in vision and sound.
A main challenge for AVE localization is to correctly distinguish the event category.

Specifically, for each data batch during training process, we compute the Euclidean distance between the video samples.
Taking a video as an anchor, we select the \emph{positive} sample identified with the same category and the largest distance.
Similarly, we select \emph{negative} videos that have different event categories from the anchor and the top-$K$ closest distances,
where $K$ is a hyper-parameter controlling the number of \emph{negative} samples.
The model is expected to correctly recognize those similar but hard to learn samples: the positive sample should gather around the anchor video while the negative ones should be pushed farther.

To this end, the contrastive objective for video-level positive sample activation can be formulated as,
\begin{equation}\label{eq:loss_vcon}
    \mathcal{L}_{\text{vpsa}} = \text{max}(0, d(\overline{\bm{f}}^a, \overline{\bm{f}}^p) - \frac{1}{K}\sum_{k=1}^{K}d(\overline{\bm{f}}^{a}, \overline{\bm{f}}_k^n) + \theta),
\end{equation}
where $d(\cdot, \cdot)$ computes the Euclidean distance between the $\ell_2$ normalized vectors, $\overline{\bm{f}}^a$ is the feature vector of an anchor video, obtained by averaging feature of video segments $\bm{f}$ (Eq.~\ref{eq:av_fusion}) along the temporal dimension, $\overline{\bm{f}}^p$ and $\overline{\bm{f}}^n$ are the features of positive and negative samples.
$K$ controls the number of the negative samples, $\theta$ denotes the minimum margin that the positive and negative samples should maintain.

\vspace{7mm}
\subsection{Classification}\label{sec:classification}
The fused audio-visual feature $\bm{f} \in \mathbb{R}^{T \times d_l}$
is send to the classifier for prediction. We detail the classifier and the objective function for the fully and weakly supervised settings below.
\subsubsection{ Classifier}\label{sec:classification}
For the fully supervised setting, as shown in Fig.~\ref{fig:system_flow}, the fused feature is further processed by two FC layers.
The classifier prediction $\bm{o}^{\text{fully}}\in \mathbb{R}^{T \times C}$ can be obtained through a softmax function.

For the weakly supervised setting, different from existing methods~\cite{lin2019dual, tian2018audio, xuan2020cross},
we add a weighting branch on the fully supervised classification module (Fig.~\ref{fig:system_flow}). It is essentially another FC layer that enables the model to further capture the differences between synchronized audio-visual pairs.
This process is summarized below,
\begin{equation}
\left\{\begin{split}
   & \bm{f}^h = {\bm{f}\bm{W}_4^{\text{weak}} \bm{W}_5^{\text{weak}}}, \\
   & \bm{\phi} = \sigma(\bm{f}^h \bm{W}_6^{\text{weak}} ), \\
   & \bm{o}^{\text{weak}} = s(f_{\text{avg}}(\bm{f}^h \odot \bm{\Phi})),
\end{split}\right.
\label{eq:o^weak}
\end{equation}
where $\bm{W}_4^{\text{weak}} \in \mathbb{R}^{d_l \times d_h}$, $\bm{W}_5^{\text{weak}} \in \mathbb{R}^{d_h \times C}$,
$\bm{W}_6^{\text{weak}} \in \mathbb{R}^{C\times 1} $ are learnable parameters in the FC layers, and $\bm{f}^h \in \mathbb{R}^{T \times C}$.
$\sigma$ and $s$ denote the sigmoid and softmax operators, respectively.
$\bm{\phi} \in \mathbb{R}^{T \times 1}$ weighs the importance of the temporal video segments, and $\bm{\Phi} \in \mathbb{R}^{T\times C}$ is obtained by duplicating $\bm{\phi}$ for
$C$ times. $\odot$ is the element-wise multiplication,
$f_{\text{avg}}$ is the average operation along the temporal dimension.
The final prediction $\bm{o}^{\text{weak}} \in \mathbb{R}^{1 \times C}$.

\subsubsection{Objective function}\label{sec:objective_function}
\textbf{Fully supervised setting.} Given the network output $\mathbf{o}^{\text{fully}}$ and ground truth $\mathbf{Y}^{\text{fully}}$, we adapt the cross entropy (CE) loss
as the basic objective function, written as,
\begin{equation}
   \mathcal{L}_{\text{ce}} = -\frac {1}{TC}  \sum_{t=1}^{T} \sum_{c=1}^{C} \bm{Y}_{tc}^{\text{fully}} {\log(\bm{O}_{tc}^{\text{fully}})}.
   \label{eq:softmax}
\end{equation}
Recall that each row of $\bm{Y}^{\text{fully}}$ contains a one-hot event label vector, describing the category of each video segment (synchronized audio-visual pair).
As such, this classification loss allows the network to predict which \emph{event category} a video segment contains.

Apart from the CE loss, we propose a new loss item, named audio-visual pair similarity loss based on the PSP $\mathcal{L}_{\text{avpsp}}$. In principle, it asks the network to produce similar features for a pair of audio and visual components if the pair \emph{contains an event} (contrasting from background) during PSP.
Specifically, for a video composed of $T$ segments, we define label vector $\bm{G} = { \{g_t | g_t \in \{ 0, 1\}, t=1,2,...,T\} \in \mathbb{R}^{1 \times T}} $,
where $g_t$ represents whether the $t^{\text{th}}$ segment is an event or background. Next, $\ell_1$ normalization is performed on $\bm{G}$.
We then compute the $\ell_1$ normalized similarity vector $\bm{S} \in \mathbb{R}^{1 \times T}$ between the visual and audio features
\begin{equation}
   \begin{split}
      \bm{S} &= \frac {\bm{v}^{\text{psp}} \odot \bm{a}^{\text{psp}}} {\left \|  \bm{v}^{\text{psp}} \odot \bm{a}^{\text{psp}} \right\|_1},
   \end{split}
   \label{eq:similarity}
\end{equation}
where $\|\cdot\|_1$ calculates the $\ell_1$ norm of a vector. The proposed loss $\mathcal{L}_{\text{avpsp}}$ is then written as,
\begin{equation}\label{eq:avps}
\begin{split}
   \mathcal{L}_{\text{avpsp}} &= \mathcal{L}_{\text{MSE}}(\bm{S}, \bm{G}),
\end{split}
\end{equation}
where $\mathcal{L}_{\text{MSE}}(\cdot, \cdot)$ computes the mean squared error between two vectors.

Combining Eq. \ref{eq:avps} and Eq. \ref{eq:softmax}, the objective function for fully-supervised setting $\mathcal{L}_{\text{fully}}$ can be computed by:
\begin{equation}
    \mathcal{L}_{\text{fully}} = \mathcal{L}_{\text{ce}} 
    + \lambda_1{\mathcal{L}_{\text{avpsp}}}.
   \label{eq:fully_loss}
\end{equation}

When refining the fused features with PSA$_S$ and PSA$_V$ jointly, the overall objective function $\mathcal{L}_{\text{fully}}^{r}$ can be computed by,
 \begin{equation}
     \mathcal{L}_{\text{fully}}^{r} = 
     \mathcal{L}_{\text{fully}} + 
     \lambda_2{\mathcal{L}_{\text{spsa}}} + 
     \lambda_3{\mathcal{L}_{\text{vpsa}}},
 \label{eq:rf_loss}
 \end{equation}
where $\lambda_1$, $\lambda_2$, $\lambda_3$ are hyper-parameters to balance the losses.
The PSA$_S$ and PSA$_V$ can also be added to the vanilla PSP separately and such manner is slightly superior than joint training, we will discuss these two strategies in Sec. 6.4.1.

\textbf{Weakly supervised setting.} For this setting,
following the practice in ~\cite{lin2019dual, xu2020MM}, we adapt the binary cross entropy (BCE) loss as the basic classification loss, formulated as,
\begin{equation}
   \mathcal{L}_{\text{weak}} = \mathcal{L}_{\text{BCE}}(\bm{o}^{\text{weak}}, \bm{Y}^{\text{weak}}).
   \label{eq:weak_bce_loss}
\end{equation}
It is worth mentioning that the segment-level event label should be known to distinguish the positive and negative segments in the video,
thus, PSA$_S$ is not applicable for weakly supervised AVE localization.
The PSA$_V$ can be used in both fully and weakly supervised settings since both of them provide the video-level event label. When combined with the loss item of PSA$_V$, the overall objective can be written as,
 \begin{equation}
   \mathcal{L}_{\text{weak}}^{r} = \mathcal{L}_{\text{weak}} + \lambda_4{\mathcal{L}_{\text{vpsa}}},
    \label{eq:rw_loss}
 \end{equation}
where $\lambda_4$ is a balance weight.

\section{Discussion}\label{sec:discussion}
\textbf{Detailed examination and meanings of $\bm{v}^{\text{pos}}$ and $\bm{a}^{\text{pos}}$.}
The computation of $\bm{v}^{\text{pos}}$ ($\bm{a}^{\text{pos}}$) is shown in Eq. \ref{eq:3}.
Take $\bm{v}^{\text{pos}}$ for example, the $i^{\text{th}}$ row $\bm{v}_{i}^{\text{pos}}$ is the weighted sum of the visual feature $\bm{v}_j^{\text{lstm}} (j=1,2,...,T)$ after linear transformation.
Here the weight, denoted as $\bm{\gamma}_{i}^{\text{av}}$, is exactly the similarity between the audio feature $\bm{a}_i$ and features of all the visual components.
Note that some elements of $\bm{\gamma}_{i}^{\text{av}}$ are zeros since the negative and weak connections are pruned during PSP, so $\bm{v}_i^{\text{pos}}$ is the aggregation result of  those {\em positive} visual features which 
are most relevant to $\bm{a}_i$.

\textbf{Physical meanings of $\bm{v}^{\text{psp}}$ and $\bm{a}^{\text{psp}}$.}
Take $\bm{a}^{\text{psp}}$ for example. From Eq. \ref{eq:3}, we find that $\bm{a}^{\text{psp}}$ is composed of two features: the original audio feature $\bm{a}^{\text{lstm}}$ and the aggregation of positive visual features $\bm{v}^{\text{pos}}$.
As discussed above, those positive visual features have large audio-visual similarity values, \ie, small vector angles and similar vector directions.
Therefore, after being added to $\bm{v}^{\text{pos}}$, the magnitude and direction of vectors representing original audio feature $\bm{a}^{\text{lstm}}$ will be changed to reflect that during training. Such an adjustment in the distribution of audio representation can be verified by the visualization results in Fig. \ref{fig:feature_distribution}.

\textbf{Why an additional FC layer in the weakly supervised setting?} When fully supervised, clear supervision is known for each segment.
For the weakly supervised setting, both the ground truth label $\bm{Y}^{\text{weak}} \in \mathbb{R}^{1 \times C}$ and the prediction $\bm{o}^{\text{weak}}\in \mathbb{R}^{1 \times C}$ are obtained through an average pooling operation along the temporal dimension.
Without knowing the supervision of each segment, the baseline approach considers all temporal video segments to have similar weights when calculating the loss.
It makes it harder for the model to focus on video segments that contain an event.
In our design, through the sigmoid activation function, 
we obtain the weights of temporal video segments.
As such, our model can better distinguish these temporal sequences and thus help locate which segments contain an event.

\textbf{Implications of $\mathcal{L}_{\text{avpsp}}$.
}
As shown in Eq. \ref{eq:softmax}, the classification loss $\mathcal{L}_{\text{ce}}$ prompts the model to calculate the loss between the output probabilities and the ground truth label.
In comparison, $\mathcal{L}_{\text{avpsp}}$ allows the network to be aware of \emph{whether an event exists in an audio-visual pair} (pair-level contrasting).
Specifically, if $g_t$ is equal to $1$, the synchronized audio-visual feature should have a higher similarity, and otherwise lower.
Therefore, for an audio (visual) component, $\mathcal{L}_{\text{avpsp}}$ provides another auxiliary constraint so that the model can better select the most relevant visual (audio) components for feature aggregation during PSP.
Note that $\mathcal{L}_{\text{avpsp}}$ cannot be adapted to the weakly supervised setting, where the label $g_t$ of each segment is unknown. To summarize, $\mathcal{L}_{\text{ce}}$ and $\mathcal{L}_{\text{avpsp}}$ serve as strong supervisions, especially in the fully supervised setting.

\textbf{Implications of $\mathcal{L}_{\text{spsa}}$ and $\mathcal{L}_{\text{vpsa}}$}.
The design of $\mathcal{L}_{\text{spsa}}$ in Eq.~\ref{eq:loss_scon} and $\mathcal{L}_{\text{vpsa}}$ in Eq.~\ref{eq:loss_vcon} are intended to further advance the audio-visual representation learning. $\mathcal{L}_{\text{spsa}}$ allows the network to be aware of \emph{whether an event exists in a segment unit}, and $\mathcal{L}_{\text{vpsa}}$ allows the network to be aware of \emph{whether an exact event category exists in a video}.
Here, the network is enforced to learn the discriminative representation capability with the most relevant segments and the same category samples. 
In fact, these two losses can be regarded as the soft supervisions since they are controlled by the hyper-parameters (\ie, $\eta$ for $\mathcal{L}_{\text{spsa}}$, $K$ and $\theta$ for $\mathcal{L}_{\text{vpsa}}$). The positive segment and video samples respectively activated by the PSA$_S$ ($\mathcal{L}_{\text{spsa}}$) and PSA$_V$ ($\mathcal{L}_{\text{vpsa}}$) are beneficial for the classifier training and this can be confirmed by the visualization examples shown in Figs.~\ref{fig:PSP_PSPCL_a},~\ref{fig:PSP_PSPCL_b}. 
To summarize, $\mathcal{L}_{\text{avpsp}}$ (PSP) and $\mathcal{L}_{\text{spsa}}$ (PSA$_S$) exploit the positive clues of intra-video correlation (using pair and segment event labels), $\mathcal{L}_{\text{vpsa}}$ (PSA$_V$) focuses on the positive clues of inter-video correlation (using video category label). As the same reason for $\mathcal{L}_{\text{avpsp}}$, $\mathcal{L}_{\text{spsa}}$ is used for fully supervised setting while $\mathcal{L}_{\text{vpsa}}$ is not limited to this constraint.

\section{Experiment}
\label{sec:experiment}
\subsection{Experimental setup}\label{sec:expriment_setup}
\textbf{Dataset.} 
\textbf{(1) AVE dataset~\cite{tian2018audio}}. Following the existing works \cite{lin2019dual, tian2018audio, xu2020MM, xuan2020cross}, we use the public AVE dataset for localization.
This dataset contains 4,143 videos,
which cover various real-life scenes and can be divided into 28 event categories, \eg, church bell, male speech, acoustic guitar, and dog barking.
Each video sample is evenly partitioned into 10 segments, and the duration of each segment is one-second.
The audio-visual event boundary on the segment level and the event category on the video level are provided. Keeping consistent with prior work, 3,339 videos are used for training, while both the validation and test set contains 402 videos.
\textbf{(2) VGGSound-AVEL100k dataset}. We construct a new large-scale VGGSound-AVEL100k datatset for AVEL task, in which the videos are sampled from VGGSound~\cite{chen2020vggsound}.
VGGSound-AVEL100k contains 101,072 videos that spans 141 audio-visual event categories covering more scenes in real-life that do not appear in the AVE dataset, such as motorboat, electric shaver, sharpen knife, \etc.
The ratio of train/validation/test split percentages are set as 60/20/20\footnote{The large-scale VGGSound-AVEL100k dataset for AVEL task is available at \href{https://drive.google.com/drive/folders/1en1dks1GYiGaDS9Ar-QtJmmyoOdzEsQj?usp=sharing}{\emph{https://drive.google.com/drive/folders/1en1dks1GYiGaDS9Ar-QtJmmyoOdzEsQj?usp=sharing}}. We give more details in Appendix.~\ref{sec:100k_details}
}.
\textbf{(3) LLP dataset~\cite{tian2020avvp}.} It is collected for a more challenging audio-visual video parsing (AVVP) task where only video-level labels are given. It contains 11,849 videos and each video in this dataset contains multiple audio and visual events. The AVVP task requires to predict what events happen in both audio and visual tracks, separately.
We extend the proposed CPSP in the weakly supervised setting to this task to evaluate the generalization ability of our method.

\textbf{Evaluation metric.} The category label of each segment is predicted in both fully and weakly
supervised settings. Following~\cite{lin2019dual, tian2018audio,xu2020MM,xuan2020cross}, we adopt
the classification accuracy of each segment as the evaluation metric.

\textbf{Training procedure and configuration.}
We have to deal with all the video data of different types (as shown in Fig.~\ref{fig:contrastive_strategy}). For convenience, we use $\mathcal{D}_{\text{bg}}$ to denote this type of video that contains both AVE and background segments (Fig.~\ref{fig:subset_bg}) and use $\mathcal{D}_{\text{ae}}$ to represent another type of video contains pure AVE segments belonging to a certain event category (Fig.~\ref{fig:subset_ae}). We initialize the proposed localization system (Fig.~\ref{fig:system_flow}) with the objective function $\mathcal{L}_{\text{fully}}$ (Eq.~\ref{eq:fully_loss}) and $\mathcal{L}_{\text{weak}}$ (Eq.~\ref{eq:weak_bce_loss}) on the dataset benchmark ($\mathcal{D}_{\text{bg}}$ \& $\mathcal{D}_{\text{ae}}$). Then we further refine the audio-visual features with PSA$_S$ on subset $\mathcal{D}_{\text{bg}}$
and PSA$_V$ on subset $\mathcal{D}_{\text{ae}}$ in Eqs.~\ref{eq:loss_scon} and \ref{eq:loss_vcon}, respectively; their corresponding usages (objective functions) for fully and weakly supervised settings are introduced in Eqs.~\ref{eq:rf_loss} and \ref{eq:rw_loss}.
More related details are discussed in Sec.~\ref{sec:evaluation_PSA}. We abbreviate the initialized network as {\textbf {PSP}}, and the refined network as {\textbf {CPSP}} in the following experiment evaluation.
The parameters are tuned on the validation set with the final model is tested on a held-out test set. The results are reported in Sec.~\ref{sec:quanly} and ~\ref{sec:quanti2}.

\textbf{Implementation details.}
(1) \emph{Visual feature extractor.}
For fair comparison, we use the VGG-19~\cite{simonyan2014very} pretrained on ImageNet~\cite{krizhevsky2017imagenet} to extract the visual features.
Specifically, 16 frames are sampled from each one-second video segment.
We extract the visual feature map for each frame from the \emph{pool-5} layer in VGG-19 with the size of $7\times7\times512 $ and then use the average map as
the visual feature for this segment.
(2) \emph{Audio feature extractor.}
For audio features, we first process the raw audio into log-mel spectrograms and then use the VGGish, a VGG-like network~\cite{hershey2017cnn}
pretrained on AudioSet~\cite{gemmeke2017audio}, to extract the acoustic feature with the dimension of 128.
(3) \emph{Hyper-parameter settings.}
The temperature parameter $\eta$ in Eq.~\ref{eq:loss_scon} is set to 0.1.
Impacts of the number of negative samples $K$ and the margin $\theta$ in Eq.~\ref{eq:loss_vcon} are discussed in Sec.~\ref{sec:evaluation_PSA}.
The weights of $\lambda_1$ in Eq. \ref{eq:fully_loss} and $\lambda_2$, $\lambda_3$ in Eq.~\ref{eq:rf_loss} is empirically set to 100, 0.01, 1, respectively.
$\lambda_4$ in Eq.~\ref{eq:rw_loss} is set to 0.005. 
These hyper-parameters remain the same on the AVE and the VGGSound-AVEL100k datasets in our experiments. In addition, the batch size in our experiments is set to 128. We use Adam~\cite{kingma2014Adam} as the optimizer, and dropout technique is used in all the linear layers  (Fig.~\ref{fig:system_flow}) with the drop rate set to 0.1.
As for the experiments on LLP dataset for video parsing, the batch size is set to 16 same as baseline method HAN~\cite{tian2020avvp}, more implementation details are introduced in Sec.~\ref{sec:avvp}.

\begin{table}[t]
   \caption{Comparison with the state-of-the-art methods under both the fully and weakly supervised settings. We report the accuracy(\%) measured on the AVE and the VGGSound-AVEL100k datasets. * indicates the number is reproduced by us.}
   \begin{center}
  \setlength{\tabcolsep}{3.5mm}
   \begin{tabular}{l|c|c|c|c}
   \toprule[0.8pt]
   \multirow{2}{*}{Method}  & \multicolumn{2}{c|}{AVE} & \multicolumn{2}{c}{VGGSound-AVEL100k} \\ \cline{2-5}
   & fully & weakly & fully & weakly \\ \hline
   AVEL~\cite{tian2018audio}      & 68.6  & 66.7  & 55.7*  & 46.2* \\
   AVSDN~\cite{lin2019dual}       & 72.6* & 67.3* & -  & - \\
   CMAN~\cite{xuan2020cross}      & 73.3* & 70.4* & - & - \\
   DAM~\cite{wu2019dual}          & 74.5  & - & -  & -  \\
   AVRB~\cite{ramaswamy2020see}   & 74.8  & 68.9 & -  & -\\
   AVIN~\cite{ramaswamy2020makes} & 75.2  & 69.4 & -  & - \\
   RFJCA~\cite{Duan_2021_WACV}    & 76.2  & - & -   & -  \\
   AVT~\cite{Lin_2020_ACCV}       & 76.8   & 70.2 & -  & - \\
   CMRA~\cite{xu2020MM}           & 77.4   & 72.9 & 57.1* & 46.8*\\
   MPN~\cite{yu2021mpn}           & 77.6  & 72.0 & - &  -\\ \hline
   PSP~\cite{zhou2021positive}(Ours)    & 77.8 & 73.5 & 58.3   & 47.4 \\
   CPSP(Ours)          & {\bf 78.6} & {\bf 74.2} & \textbf{59.9}   & \textbf{48.4} \\ \bottomrule[0.8pt]
   \end{tabular}
   \end{center}
   \label{table_4}
\end{table}

\subsection{Comparison with state of the arts}
We compare our method with the state of the arts in Table~\ref{table_4} by evaluating on the AVE and the VGGSound-AVEL100k datasets. Taking the results on the AVE dataset for example, compared with the baseline method AVEL~\cite{tian2018audio}, the CPSP exceeds it by 10.0\% and 7.5\% under the fully and weakly supervised settings, respectively. Such superiority is also proved by the results shown in Table \ref{table_1}. Also, our method exceeds those SOTAs~\cite{xu2020MM, xuan2020cross, ramaswamy2020makes, ramaswamy2020see} that focus on the cross-modal feature fusion using all of the audio-visual pairs. This indicates the necessity of the positive pair selection in PSP.
CMRA~\cite{xu2020MM} has comparable performance with the PSP method, but the CPSP is superior than CMRA on both datasets in both settings. Also, the CPSP exceeds the PSP method by a large margin.
This again demonstrates the effectiveness of the PSA performing additional contrastive learning from both segment-level and video-level. 
Such advantages of the proposed CPSP method can also be observed on the large-scale VGGSound-AVEL100k dataset.
For example, the CPSP exceeds the competitive CMRA and vanilla PSP by a large margin.
We also notice that all the methods have a performance drop on the large-scale VGGSound-AVEL100k compared to the AVE dataset, we speculate VGGSound-AVEL100k contains much more videos with more event categories that makes the problem more challenging.
Nevertheless, our method is more superior and robust under all the settings, which can be attributed to our system design.

\begin{figure*}[t]
   \begin{center}
       \includegraphics[width=\textwidth]{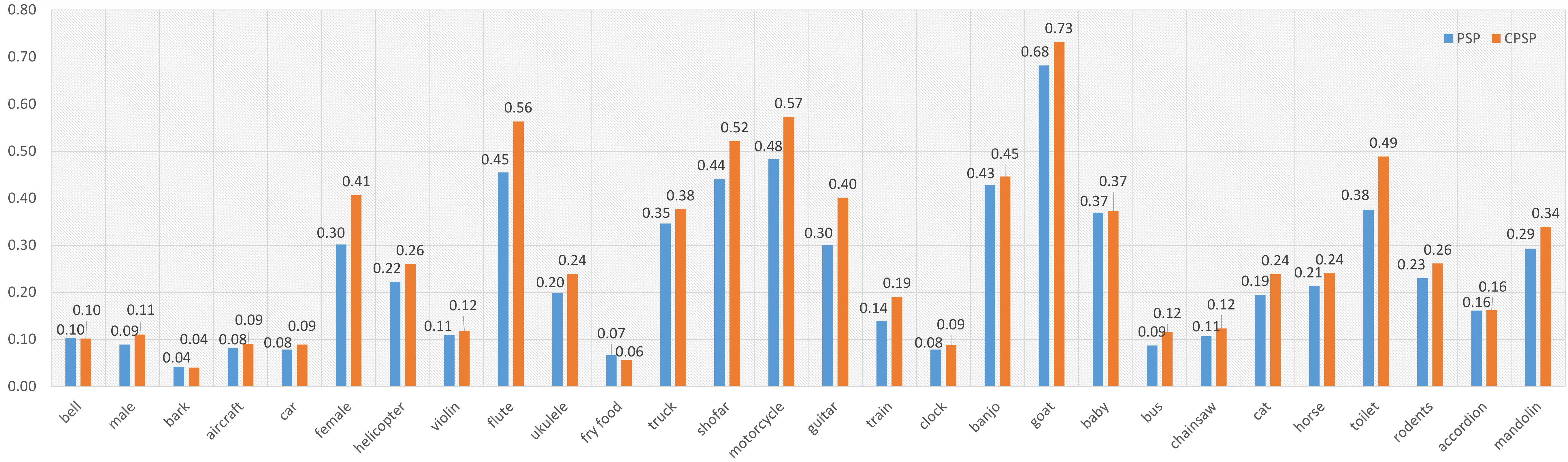}
   \end{center}
      \vspace{-0.6cm}
      \caption{Euclidean distances between the centroids of event and the background segments for each event category in the fully supervised setting. We respectively evaluate the segment features learned by the PSP and CPSP (merely w. PSA$_S$)
      in fully supervised setting.
      Larger distance of the CPSP demonstrates the benefit of PSA$_S$ helping to encode features of event and background that are easier to distinguish.
      This experiment is conducted on the AVE dataset.
      }
   \label{fig:distance_between_event_and_bgs}
  \vspace{-3mm}
\end{figure*}

\subsection{Quantitative analysis - main modules}
\label{sec:quanly}
Here we test the effects of the PSP and PSA modules. Ablation experiments are mainly conducted on the AVE dataset.
\subsubsection{Evaluation of the proposed PSP module}\label{sec:evaluation_psp}
\textbf{The effectiveness of the PSP encoding} can be verified through an ablation study in Table~\ref{table_2}. In Table \ref{table_2}, we denote the method without PSP, \ie, removing it from the localization network (Fig.~\ref{fig:system_flow}), as ``w/o PSP''.
We observe from the table that
the performance on the AVE dataset drops in both the fully supervised and weakly supervised settings significantly. Specifically, the accuracy decrease is 4.1\% (from 77.8\% to 73.7\%) and 3.3\% (from 73.5\% to 70.2\%) for the two settings, respectively. This experiment clearly validates the PSP.

\textbf{Comparison with alternative pair-level positive sample selection methods.} In our method, we emphasize that weak and negative samples are filtered out. Here, we compare this strategy with two variants: (1) all connections are used (denoted as ``ASP''); (2) only negative ones are removed, while weak connections are remained (denoted as ``WPSP''). Results are shown in Table \ref{table_2}. We have two main observations. 
First, when all samples are propagated, the accuracy of  ``ASP'' drops by 1.9\% and 2.3\% on the fully and weakly supervised settings, respectively. This shows that it is essential to have a selection process before feature aggregation instead of utilizing all the connections.
Second, although we merely remove the negative connections (\ie, with a similarity value below $\tau = 0$), the system of  ``WPSP'' is inferior to the full method. Specifically, the classification accuracy decreases by 1.8\% and 2.3\% under the fully and weakly supervised settings, which validates the effectiveness of filtering out the negative connections.

\begin{table}[t]
    \caption{Ablation studies of the proposed PSP, measured by accuracy(\%) on the AVE dataset. ``w/o'' denotes ``without''. ``ASP'' means retaining all connections ($\tau = -\infty$), while ``WPSP'' uses the weak and positive ones ($\tau = 0$). ``SAPSP'' represents adding self-attention to the feature extractor.}
   \setlength{\tabcolsep}{4mm}
   \begin{center}
   \begin{tabular}{l|c|c}
   \toprule[0.8pt]
   Method & Fully-supervised & Weakly-supervised \\
   \hline
   w/o PSP & 73.7 & 70.2 \\
   ASP  &  75.9 & 71.2\\
   WPSP & 76.0 & 71.2 \\
   SAPSP & 75.4 & 70.8 \\\hline
   PSP (ours) & {\bf 77.8} & {\bf 73.5} \\
   \bottomrule[0.8pt]
   \end{tabular}
   \end{center}
   \label{table_2}
\vspace{-3mm}
\end{table}

\begin{table}[t]
    \caption{Impact of various values of $\tau$ on the system accuracy evaluated on the AVE dataset. Results on the two settings are shown.}
   \setlength{\tabcolsep}{1.95mm}
   \small
   \begin{center}
   \begin{tabular}{l|c|c|c|c|c}
   \toprule[0.8pt]
   $\tau$ & 0 & 0.025 & 0.075 & 0.095 & 0.115 \\
   \hline
   Fully-supervised & 76.0 & 76.1 & 75.3 & {\bf 77.8} & 76.6 \\
   Weakly-supervised & 71.2 & 71.7 & 70.4 & {\bf 73.5} & 72.8  \\
   \bottomrule[0.8pt]
   \end{tabular}
   \end{center}
   \label{table_3}
\vspace{-5mm}
\end{table}

\begin{figure*}[htbp]
   \begin{center}
       \includegraphics[width=\textwidth]{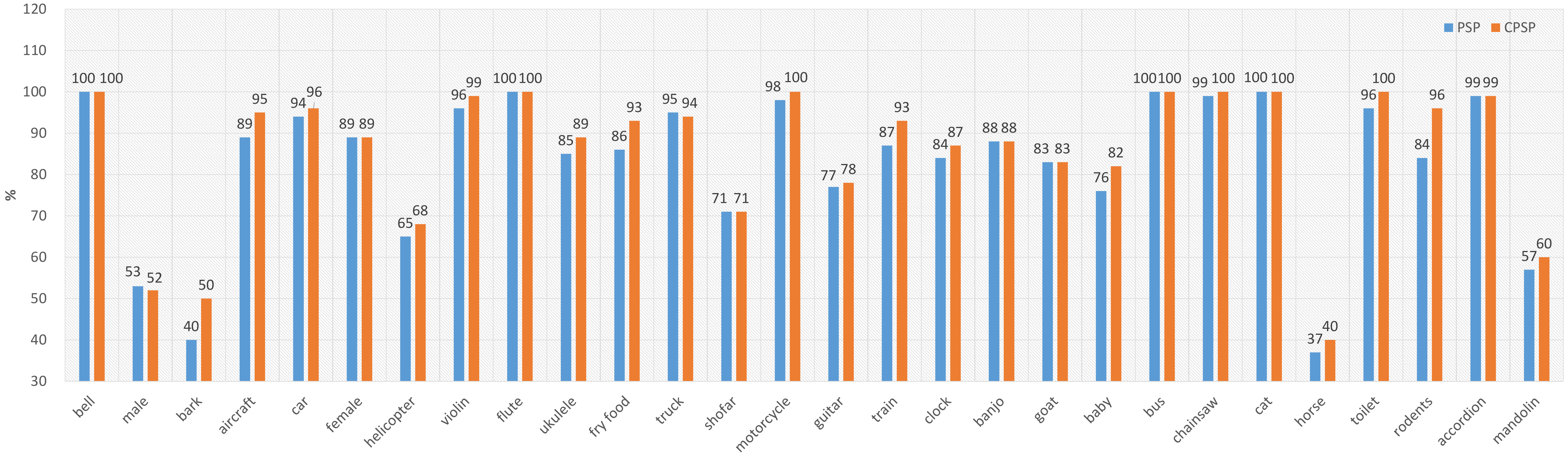}
   \end{center}
      \vspace{-0.6cm}
      \caption{Classification accuracy for videos containing no background segments in the fully supervised setting.
      Compared with the PSP, the CPSP (merely w. PSA$_V$)
      has overall superior performance in predicting the correct event categories.
      This experiment is conducted on the AVE dataset.
      }
   \label{fig:acc_for_ae_videos}
  \vspace{-0.5cm}
\end{figure*}

\textbf{Sensitivity to hyper-parameter $\tau$.} The selection process is controlled by $\tau$, determining how many connections will be cut off.
Its influence on the system accuracy is shown in Table~\ref{table_3}. We observe that the accuracy generally remains stable when $\tau$ varies between 0 and 0.115 and that the highest accuracy is achieved when $\tau=0.095$.
For different videos, the proportion of segments that are cut off highly depends on the video itself. If the whole video contains the same event of interest, it is likely that most will be retained in training; if a video contains lots of background, the same threshold will cut off more of its content.
Such a connection pruning (\ie, positive pair selection) process
in PSP can be clearly observed from the visualization example in Fig.~\ref{fig:localization_example}.

\textbf{Comparison with adding self-attention \cite{vaswani2017attention} to the feature extractor.}
Self-attention~\cite{vaswani2017attention} is widely used in existing methods
~\cite{tian2020avvp, wu2019dual, xu2020MM, xuan2020cross} to capture relationships within single modality.
To explore whether it is useful in our system, we add a self-attention module before the Bi-LSTMs in the feature extractor module and denote it as the ``SAPSP'' method.
As shown in Table \ref{table_2}, the performance surprisingly decreases by
2.4\% and 2.7\%
under fully and weakly supervised settings, respectively.
This indicates that in our system, it is not required to add additional intra-modal verification through self-attention before the PSP module.
We speculate that the PSP is sufficient to describe the cross-modality while implicitly reveals the intra-modality correlations.

\vspace{3mm}
\subsubsection{Evaluation of the proposed PSA module}\label{sec:evaluation_PSA}
\textbf{Effectiveness of the PSA$_S$.} PSA$_S$ is expected to constraint the model to learn consistent features for the video segments containing the same event, while possibly be distinguishable from the background segments. We reflect its effect by the distance of centroids of the event and background segments in feature space.
For videos in the dataset, we encode the segment features by the PSP and CPSP (merely equipped with PSA$_S$ for fair comparison), respectively.
Specifically, for each same event category, we filter out and average the features of event and background segments respectively; we take the two obtained vectors as event and background centroids. We calculate their Euclidean distance. Results on the AVE dataset are presented in Fig.~\ref{fig:distance_between_event_and_bgs}.
We can see that the distances between event and background segments are increased in most of the categories (24 out of 28) using the CPSP method. For example, for the event of \emph{female}, \emph{guitar}, and \emph{bus}, the distances increase by around 33\%.
This verifies the benefit of the PSA$_S$ that activates the event segments such that they can be better recognized from the backgrounds.

\textbf{Effectiveness of the PSA$_V$.}
As introduced in Sec.~\ref{sec:contrastive_learning},
PSA$_V$ aims to distinguish the positive video from the top-$K$ closest but negative samples, and the Euclidean distance between the video-level representations is expected to be no less than the margin $\theta$ (Eq.~\ref{eq:loss_vcon}). Here, we test CPSP merely with
PSA$_V$ for fair comparison. 
We conduct a study on the AVE dataset to explore the impacts of parameters $K$ and $\theta$ in PSA$_V$.
First, we empirically fix the number of negatives $K$ to 4 and sample $\theta$ from \{0.2, 0.4, 0.6\}.
As shown in Table~\ref{table:parameter_VLPSR}, the performance is gradually improved as $\theta$ increases. And the best accuracy is achieved when $\theta=0.6$ for both settings (\ie, $78.31\%$ for fully supervised, $74.20\%$ for weakly supervised).
We speculate that this is a relatively large margin to better distinguish the positive and negative samples.
Next, we fix $\theta$ to 0.6 and test $K$ with values \{1, 2, 4, 6\}.
As observed from the table, $K=4$ is the optimal setup. This means four negative video samples are selected from a batch of data during training to compare with the positive one.
In this way, the model enables to simultaneously compare videos of multiple event categories at once in an online fashion. We set $K=4$ and $\theta=0.6$ for all of our other experiments on both the AVE and the VGGSound-AVEL100k dataset when conducting PSA$_V$.
It is worthy to note that the performances of almost all the setups in the CPSP exceed the results of the PSP (\ie, obtained from the case without PSA$_V$, $77.8\%$ and $73.5\%$ accuracy for fully and weakly supervised settings, respectively).

\begin{table}[t]
\begin{center}
\caption{Parameter study of the $K$ and $\theta$ in PSA$_V$. We report the performance of the CPSP under different PSA$_V$ setups in both fully and weakly supervised settings.
Experiments are conducted on the AVE dataset. The \textbf{bold-faced} results represent the optimal performance is achieved under that setup.}
\begin{tabular}{l|l|c|c}
\toprule[0.8pt]
\multicolumn{2}{l|}{Parameter setup} & Fully-supervised & Weakly-supervised \\ \hline
\multirow{3}{*}{$K=4$} & $\theta=0.2$    & 77.61 & 74.10  \\
& $\theta=0.4$    & 78.10 & 74.18  \\
& $\theta=0.6$    & \textbf{78.31} & \textbf{74.20}  \\ \hline
\multirow{4}{*}{$\theta=0.6$}    & $K=1$      & 78.20 & 74.15  \\
& $K=2$     & 78.13 & 74.15  \\
& $K=4$     & \textbf{78.31} & \textbf{74.20}  \\
& $K=6$     & 78.23 & 74.10  \\
\bottomrule[0.9pt]
\end{tabular}
\label{table:parameter_VLPSR}
\end{center}
\vspace{-5mm}
\end{table}

To clarify the effect of PSA$_V$ more clearly, here we report the classification accuracy for each event category under the fully supervised setting on subset $\mathcal{D}_{\text{ae}}$ of AVE dataset, where videos in $\mathcal{D}_{\text{ae}}$ contain no background segments (\ie. having definite video-level category).
The results are shown in Fig.~\ref{fig:acc_for_ae_videos}, the CPSP (merely equipped with PSA$_V$ here) has better performance in most event categories (26 out of 28).
This verifies that PSA$_V$ can further help to predict the accurate event category, which is contributed to the video-level contrastive learning that makes the learned features more distinguishable for videos owing to different categories.

\begin{table*}[t]
\caption{
Results on the AVE and VGGSound-AVEL100k datasets are reported.
We list details of the experimental configurations, \emph{i.e.}, the objective function (Objective), the type of video data used in the objective optimization (Data), the initialized model
(Init.),  the trained model (Return), and the learning rate (Lr).}
\vspace{-5mm}
\begin{center}
\resizebox{\textwidth}{!}{
\begin{tabular}{llllllllllc}
\toprule[0.8pt]
\multicolumn{11}{c}{Fully-supervised setting}  \\ \cline{1-11}
\multirow{2}{*}{Method} & \multicolumn{4}{c}{\multirow{1}{*}{Objective}} &\multicolumn{1}{l}{\multirow{2}{*}{Data}} &  \multicolumn{1}{l}{\multirow{2}{*}{Init.}} &   \multicolumn{1}{l}{\multirow{2}{*}{Return}}& \multicolumn{1}{l}{\multirow{2}{*}{Lr}} & \multicolumn{2}{c}{Accuracy}  \\ \cline{2-5}
\cline{10-11}
& \multicolumn{2}{c}{$\mathcal{L}_{\text{ce}}$\&$\mathcal{L}_{\text{avpsp}}$}  &$\mathcal{L}_{\text{spsa}}$ & $\mathcal{L}_{\text{vpsa}}$ & \multicolumn{1}{l}{}  & \multicolumn{1}{l}{}  & \multicolumn{1}{l}{}  & \multicolumn{1}{l}{}  & AVE & VGGSound-AVEL100k   \\ \hline
PSP & \multicolumn{2}{c}{\checkmark} & &  & $\mathcal{D}_{\text{bg}}$\& $\mathcal{D}_{\text{ae}}$ & Xavier~\cite{glorot2010xavier}  & $\mathcal{M}_{\text{fully}}^{\emph{\text{p}}}$   & $10^{-3}$  & 77.8 & 58.3   \\ \hline
CPSP$_S$  & \multicolumn{2}{c}{\checkmark}  &\checkmark & &  $\mathcal{D}_{\text{bg}}$  & $\mathcal{M}_{\text{fully}}^\emph{\text{p}}$   & $\mathcal{M}_{\text{fully}}^\emph{\text{sp}}$    & $10^{-4}$  & 78.2 & 59.6  \\
CPSP$_V$ & \multicolumn{2}{c}{\checkmark} &\multirow{1}{*}{} &\multirow{1}{*}{\checkmark}  & \multirow{1}{*}{$\mathcal{D}_{\text{ae}}$}   & $\mathcal{M}_{\text{fully}}^\emph{\text{p}}$     & $\mathcal{M}_{\text{fully}}^\emph{\text{vp}}$      & \multirow{1}{*}{$10^{-5}$}    & 78.3 & 59.8\\ \hline
CPSP(join)  & \multicolumn{2}{c}{\checkmark} &\multirow{1}{*}{\checkmark} &\multirow{1}{*}{\checkmark}  &$\mathcal{D}_{\text{bg}}$\& $\mathcal{D}_{\text{ae}}$    & $\mathcal{M}_{\text{fully}}^\emph{\text{p}}$   & $\mathcal{M}_{\text{fully}}^\emph{\text{vsp}}$  & \multirow{1}{*}{$10^{-5}$}   & 78.4 & 59.8 \\ 
CPSP(sepa)  & \multicolumn{2}{c}{\checkmark} &\multirow{1}{*}{\checkmark} &\multirow{1}{*}{\checkmark}  & \multirow{1}{*}{$\mathcal{D}_{\text{bg}} \rightarrow\mathcal{D}_{\text{ae}}$}    & $\mathcal{M}_{\text{fully}}^\emph{\text{sp}}$   & $\mathcal{M}_{\text{fully}}^\emph{\text{vsp}}$  & \multirow{1}{*}{$10^{-5}$}   & \textbf{78.6} & \textbf{59.9} \\  \hline
 \hline
\multicolumn{11}{c}{Weakly-supervised setting}  \\ 
\cline{1-11}
\multirow{2}{*}{Method} & \multicolumn{4}{c}{\multirow{1}{*}{Objective}} &\multicolumn{1}{l}{\multirow{2}{*}{Data}} &  \multicolumn{1}{l}{\multirow{2}{*}{Init.}} &   \multicolumn{1}{l}{\multirow{2}{*}{Return}}& \multicolumn{1}{l}{\multirow{2}{*}{Lr}} & \multicolumn{2}{c}{Accuracy} \\ \cline{2-5}
\cline{10-11}
 & \multicolumn{2}{c}{$\mathcal{L}_{\text{bce}}$}  & \multicolumn{2}{c}{$\mathcal{L}_{\text{vpsa}}$}  & \multicolumn{1}{l}{}  & \multicolumn{1}{l}{}  & \multicolumn{1}{l}{}  & \multicolumn{1}{l}{}  & AVE & VGGSound-AVEL100k   \\ \hline
PSP  & \multicolumn{2}{c}{\checkmark} & \multicolumn{2}{c}{}   & $\mathcal{D}_{\text{bg}}$ \& $\mathcal{D}_{\text{ae}}$ & Xavier~\cite{glorot2010xavier}    & $\mathcal{M}_{\text{weak}}^\emph{\text{p}}$    & $10^{-3}$  & 73.5  & 47.4 \\
CPSP & \multicolumn{2}{c}{\checkmark} & \multicolumn{2}{c}{\checkmark}  & $\mathcal{D}_{\text{ae}}$ & $\mathcal{M}_{\text{weak}}^\emph{\text{p}}$    & $\mathcal{M}_{\text{weak}}^\emph{\text{vp}}$   & $10^{-5}$  &\textbf{74.2} & \textbf{48.4}\\ \bottomrule[0.8pt]
\end{tabular}}
\end{center}\label{table:training_strategy}
\vspace{-5mm}
\end{table*}

\begin{table}[t]
    \caption{Method comparison on the AVE dataset under two settings. We evaluate 1) loss $\mathcal{L}_{\text{avpsp}}$ under the fully supervised setting, and 2) the weighting branch under the weakly supervised setting. The two improvements are implemented on top of our system and AVEL \cite{tian2018audio}. Under AVEL, * denotes that the number is produced by us. We use \textbf{bold} font to show the higher performance brought by our technique.}
   \setlength{\tabcolsep}{3mm}
   \begin{center}
   \begin{tabular}{l|l|c|c}
   \toprule[0.8pt]
   \multicolumn{1}{l|}{Setting} & Method     & PSP~\cite{zhou2021positive}(ours) & AVEL~\cite{tian2018audio} \\ \hline
   \multirow{2}{*}{fully} & $\mathcal{L}_{\text{ce}}$ & 76.6     & 69.8*   \\
& $\mathcal{L}_{\text{ce}} + \lambda_1\mathcal{L}_{\text{avpsp}}$ & {\bf 77.8}    & \bf 71.3*   \\ \hline
 \multirow{2}{*}{weakly} & w/o weight. branch  & 71.6 & 66.9*   \\
 & w/ weight. branch   & {\bf 73.5} & \bf 69.2*   \\ \bottomrule[0.8pt]
   \end{tabular}
   \end{center}
   \label{table_1}
\vspace{-3mm}
\end{table}

\subsubsection{Evaluation of the improvement in fully/weakly setting}\label{sec:evaluation_improvements}
\textbf{Effectiveness of the pair similarity loss $\mathcal{L}_{\text{avpsp}}$ in fully supervised setting.}
We respectively adapt $\mathcal{L}_{\text{ce}}$
and $\mathcal{L}_{\text{ce}}+\lambda_1\mathcal{L}_{\text{avpsp}}$ as the objective function and test them for model training. Two baselines are used: our PSP system and the AVEL system \cite{tian2018audio}. Results are presented in Table \ref{table_1}.
We can clearly see that $\mathcal{L}_{\text{avpsp}}$ improves the accuracy when the system is fully supervised. The improvement is 1.2\% and 1.5\% for PSP and AVEL, respectively. These results confirm the role of $\mathcal{L}_{\text{avpsp}}$ as an auxiliary restriction to help to select the positive audio-visual pairs for feature aggregation.

\textbf{Improvement from the additional FC in the weakly supervised setting.}
In the weakly supervised setting, the major difference between our classification module and traditional methods~\cite{lin2019dual, tian2018audio, xuan2020cross} consists in the weighting branch (Fig. \ref{fig:system_flow}).
To evaluate its effectiveness, we also implement this branch on top of the PSP and AVEL baselines.
The results are shown in the last two rows of Table \ref{table_1}. We find that the performance
of PSP and AVEL is improved by
1.9\% and 2.3\%,
respectively.
We argue that the additional weighting branch within the designed classification module allows the model to give different weights to the temporal sequences, thus benefiting the localization of the target video segments. 
These results confirm the effectiveness of the proposed improvements and show their robustness in other localization network. We refer readers to Sec. \ref{sec:discussion} for methodological discussions on the this technique.

\subsection{Quantitative analysis - contrastive manner in CPSP}\label{sec:quanti2}
In this subsection, we first compare the PSP (without contrastive learning) and the CPSP (with contrastive learning), and then discuss the supervised CPSP and self-supervised PSP (named SSPSP).

\subsubsection{Comparison of the PSP and CPSP}

To reveal the impact of each contrastive loss, we list five training modes in Table~\ref{table:training_strategy}:
\textcircled{1} the vanilla PSP~\cite{zhou2021positive}, 
\textcircled{2} the CPSP with sole PSA$_S$ (denoted as {\bf CPSP$_S$}), 
\textcircled{3} the CPSP with sole PSA$_V$ (denoted as {\bf CPSP$_V$}), 
\textcircled{4} the CPSP with both PSA$_S$ and PSA$_V$ by jointly training (denoted as {\bf CPSP(join)}), and 
\textcircled{5} the CPSP with both PSA$_S$ and PSA$_V$ by separately training (first PSA$_S$ then PSA$_V$, denoted as {\bf CPSP(sepa)}).

\begin{figure*}[htbp]
\begin{center}
\includegraphics[width=\textwidth]{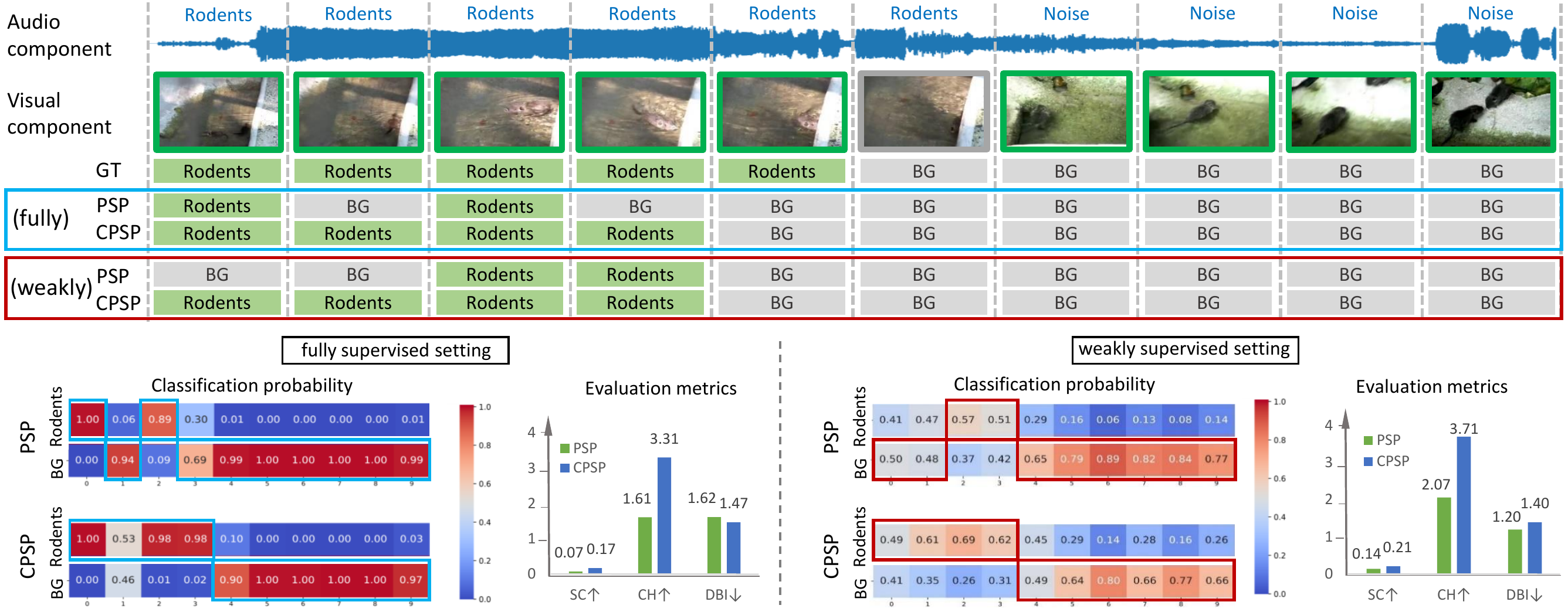}
\caption{Localization results of a qualitative example (sampling from $\mathcal{D}_{\text{bg}}$ of AVE dataset) under both fully and weakly supervised settings. In this example, the first five video segments contain the audio-visual event \emph{Rodents}.
In either setting, PSP wrongly predicts some segments as \emph{Background} (gray boxes), the CPSP method has the correct results (green boxes). The classification probability maps confirm this.
We also compare the features learned by the PSP and CPSP using the metrics aforementioned in Table~\ref{table_sc_ch_dbi}. These metrics reflect the clustering quality of features.
The results show that the features learned by the CPSP are more distinguishable.
It also verifies the segment-level event-awareness capability of the CPSP.}
\label{fig:PSP_PSPCL_a}
\end{center}
\vspace{-3mm}
\end{figure*}

As shown in Table~\ref{table:training_strategy}: (1) Compared with the vanilla PSP, the performances are improved by the CPSP with PSA$_S$ and PSA$_V$ in both fully and weakly supervised settings for both datasets. Take the VGGSound-AVEL100k dataset for example, after utilizing the PSA$_S$ or PSA$_V$ separately for the fully supervised setting, the accuracy of \textbf{CPSP$_S$} and \textbf{CPSP$_V$} increase by $1.3\%$ and $1.5\%$ (from $58.3\%$ to $59.6\%$ and $59.8\%$), respectively;
(2) When jointing the PSA$_S$ and PSA$_V$ together, the performances keep stable under both combination strategies, \ie, \textbf{CPSP(join)} and \textbf{CPSP(sepa)}.
The results of \textbf{CPSP(join)} and \textbf{CPSP(sepa)} are comparable and \textbf{CPSP(join)} performs slightly worse. We argue that it may be a little confusing for \textbf{CPSP(join)} to train with the PSA$_S$ and PSA$_V$ simultaneously (totally different contrastive goals).  
With \textbf{CPSP(sepa)}, the best accuracy can achieve $78.6\%$ and $59.9\%$ for the AVE and the VGGSound-AVEL100k datasets, respectively.
The CPSP is flexible to be utilized: with single independent PSA$_S$ or PSA$_V$ module, or with both modules under either training setup.
Anyway, these benefit from the proposed PSA by activating the model
to distinguish (1) positive and negative segments in PSA$_S$,
and (2) event categories in PSA$_V$, the model gains a more robust capability to correctly identify the event location and its categories.
This is consistent with the goal of AVE localization thus can promise better results, which can also be verified by the qualitative examples as shown in Figs.~\ref{fig:PSP_PSPCL_a},~\ref{fig:PSP_PSPCL_b}.

\begin{table}[t]
\caption{Quality analysis of the encoded features of videos in the AVE dataset. We measure three clustering metrics (SC, CH, DBI) in two ways: ``Mean" denotes evaluating two clusters (event and background) in each event category, while ``All" refers to $C-1$ clusters, where $C-1$ is the number of all the event categories except background.
}
\begin{center}
\setlength{\tabcolsep}{2.5mm}
\begin{tabular}{l|c|c|c|c|c|c}
\toprule[0.8pt]
\multicolumn{1}{c|}{\multirow{2}{*}{Method}} & \multicolumn{2}{c|}{SC $\uparrow$} & \multicolumn{2}{c|}{CH $\uparrow$} & \multicolumn{2}{c}{DBI $\downarrow$} \\ \cline{2-7}
\multicolumn{1}{c|}{} & Mean & All & Mean & All & Mean & All \\ \hline
PSP & 0.17 & 0.19  & 16.64 & 194.12 & 1.62 & 2.07\\
CPSP & {\bf 0.21} & {\bf 0.22}  & {\bf 18.29} & {\bf 195.51} & {\bf 1.54} & {\bf 1.95} \\ \bottomrule[0.8pt]
\end{tabular}
\end{center}
\label{table_sc_ch_dbi}
\vspace{-3mm}
\end{table}

\begin{table*}[t]
   \caption{Comparison with the baseline methods on the test set of LLP dataset.
   $\dag$ denotes the results reported in the paper HAN~\cite{tian2020avvp}.}
   \vspace{-3mm}
   \begin{center}
\begin{threeparttable}
   \begin{tabular}{lp{0.6cm}<{\centering}p{0.6cm}<{\centering}p{0.6cm}<{\centering}p{1.4cm}<{\centering}p{1.5cm}<{\centering}p{0.6cm}<{\centering}p{0.6cm}<{\centering}p{0.6cm}<{\centering}p{1.4cm}<{\centering}p{1.5cm}<{\centering}}
   \toprule[0.8pt]
   \multirow{3}{*}{Method} & \multicolumn{5}{c}{Segment-level} & \multicolumn{5}{c}{Event-level}\\ \cmidrule(r){2-6} \cmidrule(r){7-11}
   & A & V & AV & Type@AV & Event@AV & A & V & AV & Type@AV & Event@AV\\ \midrule
  AVEL~\cite{tian2018audio} $\dag$ & 47.2 & 37.1 & 35.4 & 39.9 & 41.6 & 40.4 & 34.7 & 31.6 & 35.5 & 36.5 \\
  AVSDN~\cite{lin2019dual} $\dag$ & 47.8 & 52.0 & 37.1 & 45.7 & 50.8 & 34.1 & 46.3 & 26.5 & 35.6 & 37.7 \\
  HAN~\cite{tian2020avvp} & \textbf{60.1} & 52.9 & 48.9 & 54.0 & 55.4 & 51.3 & 48.9 & 43.0 & 47.7 & 48.0 \\ \midrule 
  PSP~\cite{zhou2021positive} & 54.2 & 54.7 & 48.3 & 52.4 & 52.5 & 46.8 & 50.2 & 42.8 & 46.6 & 45.6 \\
  CPSP (Ours) & 58.5 & \textbf{57.8} & \textbf{52.6} & \textbf{56.3} & \textbf{55.8} & \textbf{51.6} & \textbf{54.0} & \textbf{46.5} & \textbf{50.7} & \textbf{49.9} \\
 \bottomrule[0.8pt]
   \end{tabular}
   \end{threeparttable}
   \end{center}
   \label{table:CPSP_on_LLP}
 \vspace{-4mm}
\end{table*}

Moreover, we introduce three widely-used clustering metrics, \ie, Silhouette Coefficient (SC)~\cite{ROUSSEEUW198753}, Calinski-Harabasz Index (CH)~\cite{calinski1974}, and Davies-Bouldin Index (DBI)~\cite{DaviesB79}. These metrics validate the data clustering quality from the intra-class aggregation and inter-class separation.
Here, we use them to evaluate the event aggregation and background separation of features learned by the PSP and CPSP.
At first, we have a brief introduction: 
(1) SC~\cite{ROUSSEEUW198753} calculates the difference of intra-class and inter-class dissimilarities and divides it by the maximum value of these two dissimilarities;
larger score denotes that clusters are dense while well separated;
(2) CH~\cite{calinski1974} is the ratio of the covariance of the intra-class data to the covariance of the inter-class data; higher score means better performance;
(3) DBI~\cite{DaviesB79} represents the average similarity between clusters; a lower value indicates better separation between clusters.
Next, we measure these metrics (\ie, SC, CH, and DBI)  in two ways: as shown in Table~\ref{table_sc_ch_dbi}, the ``Mean'' denotes that we first split the data into event and background segments (\ie, 2 clusters, binary separation) in each event category, and then average the metrics over all the categories; the ``All'' refers to $C-1$ clusters, including all the different event categories except background (\ie, multi-class event separation).
In other words, we adopt ``Mean'' to measure the clustering effect with the partition of event and background, while ``All'' with the partition of all the event categories. 
At last, experiment is performed on the AVE dataset.
As observed from Table~\ref{table_sc_ch_dbi}, for all of the metrics, CPSP is better scored than PSP in any measurement method. This demonstrates that the positive features learned through CPSP are better clustered thus making it easier to distinguish the event segments from backgrounds, and also performing better for classifying videos with different categories.

\subsubsection{Comparison of the CPSP and SSPSP}
The contrastive learning is always conducted in a self-supervised manner in the audio-visual field \cite{afouras20ssl, ma2021contrastive, wu2021exploring}. Specifically, the synchronized audio-visual segment pair is regarded as a positive sample and otherwise is negative. There is a drawback that this manner will inevitably bring false negatives of the audio-visual pair depicting the same event (existing AVC) but at different timestamp.
We are curious about the effect of using such self-supervised manner in AVE localization task. 
So we introduce the self-supervised learning into our positive sample propagation, and denote it as ``SSPSP'' method.
In ``SSPSP'', all unsynchronized features (\ie, audio feature $\bm{a}^{\text{psp}}$ and visual feature $\bm{v}^{\text{psp}}$) are treated as negative instances sampling from a batch of data during training.
The corresponding contrastive objective can be written as $\mathcal{L}_{\text{ss}}$ below. 
We first train a vanilla PSP with $\mathcal{L}_{\text{fully}}$ (Eq.~\ref{eq:fully_loss}), and then inject the self-supervised learning to the PSP. The total objective function can be computed by
\begin{equation}
\left\{
\begin{gathered}
\begin{split}
    & \mathcal{L}_{\text{sspsp}} =
     \mathcal{L}_{\text{ce}} +
     \lambda_\text{2}^{'}{\mathcal{L}_{\text{ss}}},\\
    & \mathcal{L}_{\text{ss}} =
    - \frac{{1}}{N^b T}\! \sum_{{\emph{i}}=\text{1}}^{{N^b T}}\! \log(\frac{\text{exp}(\frac{\text{sim}(\bm{a}_\emph{i}^{\text{psp}}, \bm{v}_\emph{i}^{\text{psp}})}{\eta})}{ \!\sum_{\emph{j}=\text{1}}^{{N^b T}}\!\text{exp}(\frac{\text{sim}(\bm{a}_\emph{i}^{\text{psp}}, \bm{v}_\emph{j}^{\text{psp}})}{\eta^{'}})}\!),
\end{split}
\end{gathered}
\right.
\label{eq:loss_SS}
\end{equation}
where $N^b$ denotes the number of videos in a batch and $T$ is the number of segment in each video; thus, $N^b \cdot T$ denotes the total segment number in a batch. $\text{sim}(\cdot, \cdot)$ computes the dot product of the $\ell_2$ normalized vectors (\emph{i.e.}, cosine similarity). The $i$ and $j$ are the indexes of the video segment, note that $j$ can also index the $i$-th segment. $\eta^{'}$ is a temperature parameter controlling the concentration level of feature distribution; it is set to 0.3 in our experiments. $\lambda_2^{'}$ is the weight to balance these two losses and is empirically set to 0.01.

\begin{table}[t]
   \caption{Comparison of different contrastive manners (CPSP v.s. self-supervised SSPSP) under two settings. We report the accuracy(\%) measured on the AVE and the VGGSound-AVEL100k datasets.
   }
   \vspace{-7mm}
   \begin{center}
  \setlength{\tabcolsep}{4.5mm}
   \begin{tabular}{l|cc|cc}
   \toprule[0.8pt]
   \multirow{2}{*}{Method}  & \multicolumn{2}{c|}{AVE} & \multicolumn{2}{c}{VGGSound-AVEL100k} \\ \cline{2-5}
   & fully & weakly & fully & weakly \\ \hline
   PSP     & 77.8  & 73.5  & 58.3  & 47.4 \\
   SSPSP   & 78.2  & 73.8  & 58.8  & 48.0 \\
   CPSP    & \textbf{78.6}  & \textbf{74.2}  & \textbf{59.9}  & \textbf{48.4} \\ \bottomrule[0.8pt]
   \end{tabular}
   \end{center}
   \label{table:SSPSP}
  \vspace{-0.5cm}
\end{table}

\begin{figure*}[t]
   \begin{center}
       \includegraphics[width=\textwidth]{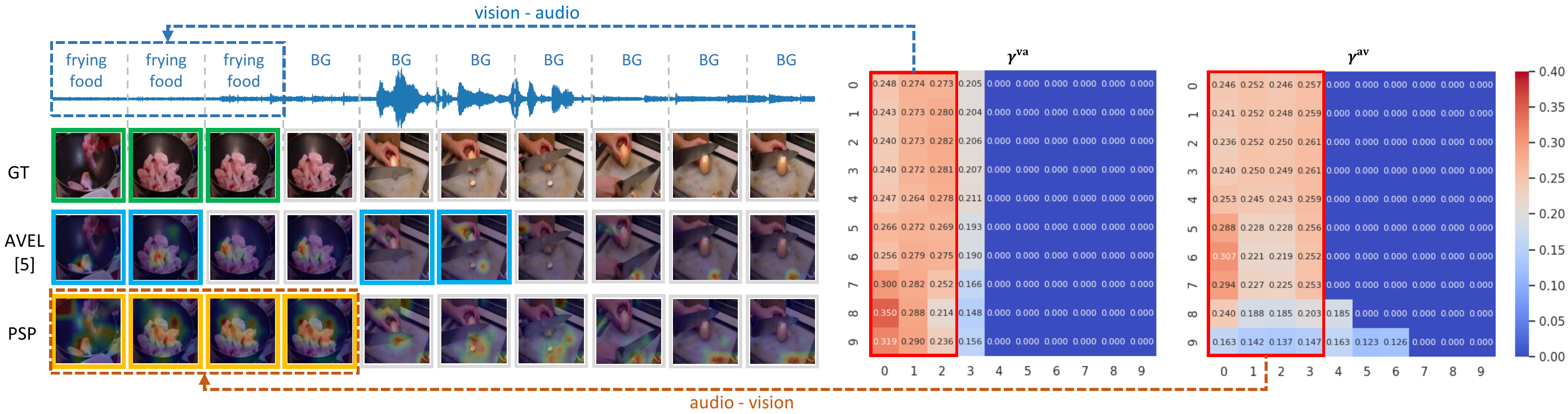}
   \end{center}
      \vspace{-0.5cm}
      \caption{A qualitative example of pair-level propagation in PSP.  
      For the video on the \textbf{left}, only the first three segments simultaneously contain the visual and audio signals of the event \emph{frying food}.
      The green boxes represent ground truth labels.
      The blue and orange boxes indicate predictions
      of AVEL~\cite{tian2018audio} and the PSP method, respectively. Besides, we visualize the attention effect on the images. It is clear that our method produces more accurate localization.
      On the \textbf{right}, we visualize the audio-visual similarity matrices $\bm{\gamma}^{\text{va}}$ and $\bm{\gamma}^{\text{av}}$ (Eq.~\ref{eq:3}) after PSP. For $\bm{\gamma}^{\text{va}}$, the x-axis and y-axis correspond to audio and visual features, respectively, and for $\bm{\gamma}^{\text{av}}$ the order is reversed. The red bounding box in $\bm{\gamma}^{\text{va}}$ shows that
      all the visual components are highly correlated with the first three audio components containing the sound of the event.
      Besides, negative and weak connections are cut off to 0 in $\bm{\gamma}^{\text{va}}$ and $\bm{\gamma}^{\text{av}}$. The color bar corresponds to the similarity strength, with red denoting high similarities and blue for low similarities.
      }
   \label{fig:localization_example}
  \vspace{-2mm}
\end{figure*}

\begin{figure*}[htbp]
\begin{center}
\setlength{\abovecaptionskip}{0.cm}
\includegraphics[width=\textwidth]{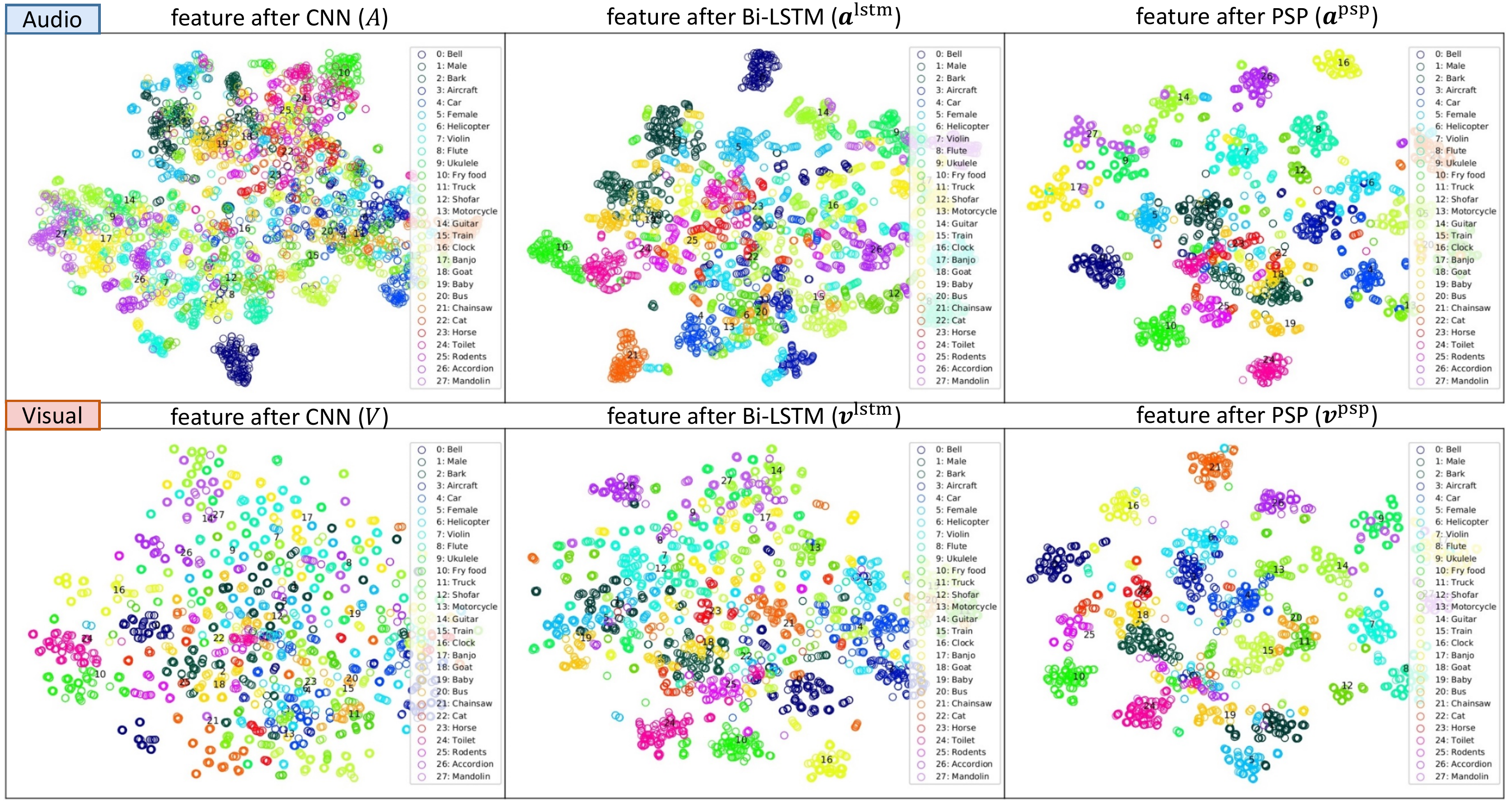}
\vspace{-3mm}
\caption{TSNE \cite{maaten2008visualizing} visualization of audio and visual feature distributions.
   The data all come from the validation set of AVE dataset under the fully supervised setting.
   (\textbf{Row 1:}) audio features.
   (\textbf{Row 2:}) visual features.
   (\textbf{Column 1:}) the CNN features.
   (\textbf{Column 2:}) features after Bi-LSTM encoding. (\textbf{Column 3:}) features after PSP encoding.
   We observe that features after PSP are much better clustered into individual classes than the Bi-LSTM and CNN features.
   Different colors represent different classes.
   Best view in color and zoom in.}
\vspace{-3mm}
\label{fig:feature_distribution}
\end{center}
\end{figure*}

The experimental result is shown in Table~\ref{table:SSPSP}, where SSPSP is conducted with the optimal experiment setup.
We can find that the performance of the SSPSP is comparable with the CPSP on the AVE dataset but is much lower on the large-scale VGGSound-AVEL100k dataset.
On VGGSound-AVEL100k, our CPSP method surpasses SSPSP by 1.1\% and 0.4\% under fully and weakly supervised settings, respectively. 
This reflects that such self-supervised contrastive method is not robust for audio-visual event localization.\footnote{We provide more experimental results and analyses in the appendix~\ref{supp_sspsp} that show the SSPSP is much more sensitive to the data distribution and training batch size, \emph{etc}.} In fact, the self-supervised SSPSP indeed ignores the semantic alignment of audio-visual pairs when constructing positive-negative samples which is vital for AVEL.
Unlike SSPSP, the proposed CPSP uses reliable audio-visual pairs to construct positive and negative samples. 
This makes the CPSP more superior.

\subsection{Quantitative analysis - generalization to AVVP task}\label{sec:avvp}
In this subsection, we extend the proposed CPSP method to a related and more challenging audio-visual video parsing (AVVP) task. 
We adopt the baseline method HAN~\cite{tian2020avvp} specifically designed for this task as the backbone and replace its core hybrid attention network for aggregating audio-visual features by the proposed PSP module.
As for the objective optimization, we keep the loss items proposed in ~\cite{tian2020avvp} and introduce the proposed video-level contrastive objective ${\mathcal{L}_{\text{vpsa}}}$ 
under the weakly-supervised labels (given only video-level labels) to adapt our CPSP model for AVVP.

Notably, since there are multiple categories of events in each video in AVVP task, there are some differences from AVEL to AVVP when constructing the positive and negative sets for contrastive learning.
Specifically, for a certain video in a batch during training, videos in the negative set can be selected from those remaining videos that have completely irrelevant event categories from it. But it is hard to select the positive samples requiring completely the same labels.
Therefore, we consider to define a co-occurrence ratio $r$ to measure the coincidence degree of event categories between two videos.
For example, given a video $Vid_a$ with event labels \{\emph{barking}, \emph{speech}\} and its pair video $Vid_b$ with labels \{\emph{speech}, \emph{music}, \emph{clapping}\}, $r$ is calculated by the proportion of co-occurrence event labels (\{\emph{speech}\}) to the total event labels of $Vid_a$ (\{\emph{barking}, \emph{speech}\}), \ie, $r$ is 1/2.
In this way, the ratio $r$ indicates that the positive samples are expected to contain as many events as possible that are appeared in the reference video (large $r$).
We consider to set a threshold $\mu$ to construct the positive samples. 
For any pair of videos, we first compute the ratio $r$ between them. 
If the $r$ is greater or equal than pre-set $\mu$ ($r \ge \mu)$, the pairwise video is selected as positive sample. 
As for the negative samples, the ratio $r$ is strictly 
equal to zero which means there are no events overlapping between the two videos.
The hyper-parameters $\mu$ and $\theta$, $K$ in Eq.~\ref{eq:loss_vcon} are empirically set to 0.6, 0.4 and 4 for AVVP, respectively.
And we provide an ablation study on the threshold $\mu$ in the appendix~\ref{supp_avvp}.

With the above setup, we train our CPSP model on LLP dataset~\cite{tian2020avvp} from scratch for AVVP.
For fair comparison, we use the same evaluation metrics as in HAN ~\cite{tian2020avvp}, referring to ``A'' and ``V''  (the F-score of audio events and visual events, respectively), ``AV'' (the F-score of audio-visual co-occurrence events, namely AVEs in AVEL task), ``Type@AV'' (the averaged result of ``A'', ``V'', and ``AV''), and ``Event@AV'' (the F-score of audio-visual events where mIoU is set to 0.5). 
From Table~\ref{table:CPSP_on_LLP}, we have three observations:
\textbf{first}, our PSP and CPSP methods surpass the baseline methods from audio-visual event localization task (AVEL~\cite{tian2018audio}, AVSDN~\cite{lin2019dual}) by a large margin for audio-visual video parsing (AVVP).
\textbf{Second}, compared to the vanilla PSP, the proposed CPSP with the video-level contrastive objective $\mathcal{L}_{\text{vpsa}}$ improves the performances significantly. For example, the metric ``Type@AV'' and ``Event@AV'' are improved by 3.9\% and 3.3\% for the segment-level, and they are  4.1\% and 4.3\% for the event-level, respectively. This again demonstrates the benefits of the proposed contrastive strategy.
\textbf{Third}, compared to the HAN~\cite{tian2020avvp} that is specially designed for AVVP, CPSP is even more superior that has better performances, especially having obvious performance superiority on ``V'' and ``AV''.
This reflects that the CPSP not only keeps a strong ability to recognize the audio-visual events but can also competently identify separate audio events and visual events.

In short, the proposed CPSP is still effective and advanced in both the AVEL and more challenging AVVP tasks providing more evidence of the model generalization.

\subsection{Qualitative analysis}\label{visualization}
\subsubsection{Visualization of the effectiveness of PSP}
\textbf{Propagated audio and visual components in PSP.}
We start by presenting an example of audio-visual event localization in Fig.~\ref{fig:localization_example}.
The event in this sample is difficult to predict because the visual images are changeable
and the audio signals are mixed with background noise.
From the figure, we have three observations.
(1). While both our method and AVEL \cite{tian2018audio} use the AVGA attention, we show that the PSP enables better attention to visual regions closely related to sound sources.
As displayed in Fig.~\ref{fig:localization_example}, for the event of {\em frying food},
our attended regions include both the frying chicken thighs and the pot, especially in the first four segments.
In comparison, AVEL only finds the thighs and very small receptive fields.
(2). Our method has a better prediction result.
AVEL seems to make decisions merely according to synchronized audio-visual segments while
our method can pay attention to visual and audio components that are at different time stamps. For example,
AVEL incorrectly regards the fifth and sixth segments as the {\em frying food} event, ignoring the third and fourth segments which are more relevant to the event.
(3). We visualize the similarity matrices $\bm{\gamma}^{\text{va}}$ and $\bm{\gamma}^{\text{av}}$ in Fig.~\ref{fig:localization_example}.
We find that only a small percentage of all the audio-visual connections are retained after PSP selection
and are closely related to the event. For example,
they tend to build strong connections (large similarity values)
between the first three audio components and the first four visual components containing the scene of "flying food", 
\ie, feature propagation merely takes place in these event-related audio-visual pairs.
Such a propagation mechanism is critical for AVE localization because
more discriminative audio-visual features can be identified with these {\em positive} connections and subsequently used in classifier training.
Through back-propagation, it allows the model to be able to attend on more sound-relevant visual regions and visual-relevant audio segments.

\begin{figure*}[t]
\begin{center}
\includegraphics[width=\textwidth]{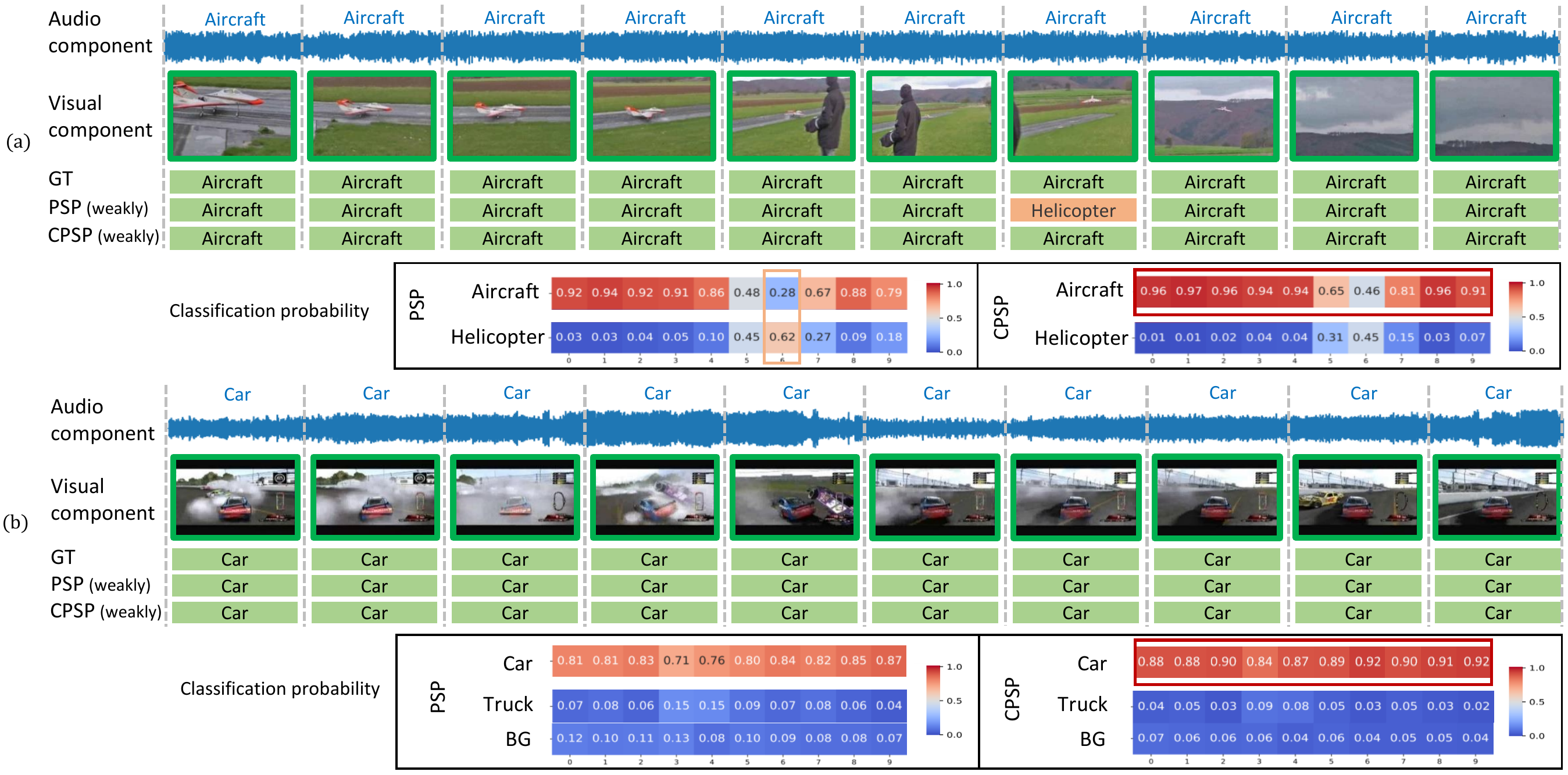}
\vspace{-5mm}
\caption{Weakly supervised setting is a more challenging setting.
We display the AVE localization results of two examples that all of the video segments contain the event (sampling from $\mathcal{D}_{\text{ae}}$ of AVE dataset).
We have two observations:
1) as shown in example (a), the PSP classifies incorrect event category in the orange box, while the CPSP provides accurate predictions in green boxes;
2) for example (b), even both the PSP and CPSP have exact predictions, the CPSP gives larger probabilities to the ground truth category.
The classification probability maps confirm this.
This indicates that the features encoded by the CPSP contain more category-aware semantics related to the ground truth thus facilitates the event category classification.}
\label{fig:PSP_PSPCL_b}
\end{center}
\vspace{-5mm}
\end{figure*}

\textbf{Feature distribution in the PSP.}
We then visualize the data distribution of features processed by different stages in our framework using TSNE~\cite{maaten2008visualizing}.
As shown in Fig.~\ref{fig:feature_distribution},
we first find that the CNN-based audio and visual features are not very well clustered.
This is because they are at a relatively low level in the network hierarchy encoding limited semantics.
Then, after Bi-LSTM, features of some categories (\eg, \emph{Rodents} and \emph{Frying food}) can be better clustered compared with the CNN features, but most are still disordered and highly mixed.
Further, after PSP, the features are much better clustered: cohesive within the same class and divergent between different classes.
This reflects that the audio-visual representations gain stronger discriminative abilities along the pipeline of our method.

\subsubsection{Visualization of the effectiveness of CPSP}
Here, we display some examples to explore the classification capability of the CPSP, where compared with the PSP, the CPSP introduces the contrastive constraints PSA$_S$ and PSA$_V$.

\textbf{Segment-level event-aware.}
First, in Fig.~\ref{fig:PSP_PSPCL_a}, we show a video example. We perform the PSP and CPSP in both fully and weakly supervised settings and report the AVE localization results.
The CPSP has more accurate predictions in both settings. Using the PSP, some segments containing the event are incorrectly classified to the background (\ie, the second and fourth segments in the fully supervised setting, the first two segments in the weakly supervised setting) while the CPSP outputs all the correct results.
We display the classification probability map to see what happens. In the fully supervised setting, the second and fourth segments are predicted to the \emph{background} with high probabilities by the PSP but this result is overturned by the CPSP. Similar phenomenon is observed in the weakly supervised setting, which performs slightly lower probabilities than full supervision
due to its poor knowledge (segment-level event labels).
We also use the evaluation metrics mentioned in Table~\ref{table_sc_ch_dbi} to test the discriminability of the segment features. As shown in the histograms, the CPSP has superior performances under all of the indicators in both settings. This reflects that positive event-aware semantics of segment features are aggregated and discriminative from the backgrounds. We speculate this is attributed to the PSA$_S$ that enforces the CPSP to learn more \emph{event-aware} semantics.

\textbf{Video-level category-aware.} We further display two examples containing no backgrounds and conduct the localization under the more challenging weakly supervised setting.
As shown in Fig.~\ref{fig:PSP_PSPCL_b}, for example (a), the PSP incorrectly classify the seventh segment to the \emph{Helicopter} (orange box; the ground truth is \emph{Aircraft}); it's even easily confused by human.CPSP takes efforts to modify the wrongly predicted result generated by the PSP (orange bounding box).
This is reflected in the classification probability map with a high probability of \emph{Aircraft} at the seventh segment.
In example (b), both the PSP and CPSP provide accurate predictions, \ie, all the segments are classified to the \emph{Car} event.
But the classification probability map tells that the CPSP gives higher probabilities to the \emph{Car} category (red bounding box), making the video segments more recognizable from the similar \emph{Truck} or \emph{background}.
These two examples demonstrate the CPSP is more \emph{category-aware} thanks to the PSA$_V$ that enables to encode features including more semantics related to the ground truth category thus distinguishing from other categories.

\section{Conclusion}\label{sec:conclusion}
For the AVE localization problem, we propose a contrastive positive sample propagation (CPSP) method that comprehensively explores three levels of positive samples for distinguishable audio-visual representation learning.
Specifically, the pair-level PSP identifies and exploits the most relevant audio and visual pairs when fusing the cross-modal features. We find that negative and weak connections, even though with small weights, have a detrimental effect on the system, and thus need to be completely removed.
The segment-level PSA$_S$ and video-level PSA$_V$ provide additional contrastive constraints to refine the features encoded by the PSP. The PSA$_S$ enforces the model to be event-aware by gathering the positive segments containing an AVE and being far away from the backgrounds. The PSA$_V$ is actually an online hard sample learning that contrasts the positive video from negatives according to the event category thus makes the model to be category-aware.
We show that such pair-level, segment-level and video-level positive sample propagation and activation method are beneficial to the classifier training.
To evaluate the model generalization ability, we collect a large-scale VGGSound-AVEL100k dataset and extend our method to a more challenging audio-visual video parsing task.
Extensive experimental results validate the effectiveness of the proposed CPSP method.
In addition, this paper covers a comprehensive study on the contrastive learning manners with different supervisions, \ie, fully-, weakly-, and self-supervised.

\ifCLASSOPTIONcompsoc
  \section*{Acknowledgments}
\else
\fi
We would like to thank the reviewers for their constructive suggestions.
This work was supported by the National Natural Science Foundation of China (72188101, 61725203, 62020106007, and 62272144), and the Major Project of Anhui Province (202203a05020011).

\ifCLASSOPTIONcaptionsoff
  \newpage
\fi

\bibliographystyle{IEEEtran}
\bibliography{IEEEabrv,IEEEtran}

\newpage
\onecolumn
\appendices

\section{
Introduction of the VGGSound-AVEL100k dataset.}\label{sec:100k_details}

In this work, the newly collected dataset provides an additional option for evaluating the generalization ability of the designed AVEL models. We wish 
it will facilitate relevant research in audio-visual community. 
VGGSound-AVEL100k contains 101,072 videos covering 141 event categories, where the video are sampled from the VGGSound~\cite{chen2020vggsound} dataset.
Category names and video number for each category are provided on our project homepage \href{https://github.com/jasongief/CPSP}{https://github.com/jasongief/CPSP}.

For data labeling, the category label can be easily obtained according to the video tags, while the temporal event labels (audio-visual correspondence labels) are manually labeled by outsourcing. Similar to the annotation process of the AVE~\cite{tian2018audio} dataset, for each video in VGGSound-AVEL100k, each annotator is required to watch through the entire video and label each video segment: if one event simultaneously occurs in both audio and visual channels, the label for the current segment is ``1'', otherwise ``0''.
To ensure the quality of labeling, 60\% of the annotated data are randomly selected and manually checked again.
In this way, we can obtain the final labels of VGGSound-AVEL100k dataset. More details such as annotation guideline, annotator information, and quality control can be seen from our website \href{https://github.com/jasongief/CPSP}{https://github.com/jasongief/CPSP}.

\section{Experiments for Audio-Visual Event Localization.}

\subsection{Multi-run experiments of CPSP}\label{supp_multi-run}

To evaluate the robustness of our method, we perform the proposed CPSP method on the AVE~\cite{tian2018audio} and the new large-scale VGGSound-AVEL100k datasets by running five times under both the fully and weakly supervised settings.
For comparison, we also reproduce some popular baseline methods - AVEL~\cite{tian2018audio}, AVSDN~\cite{lin2019dual}, CMRA~\cite{xu2020MM}, CMAN~\cite{xuan2020cross}, and PSP~\cite{zhou2021positive}.
The results are shown in Table~\ref{table:avg_result_on_AVE_VGGSound} from which we have these observations:
\textbf{first}, the top performances on AVE and VGGSound-AVEL100k during five tuns that demonstrate the generalization ability of CPSP to both small and large-scale datasets.
\textbf{Second}, in all the cases, the vanilla PSP still outperforms most of the SOTAs~\cite{tian2018audio, lin2019dual, xuan2020cross}.
CMRA~\cite{xu2020MM} is competitive with the PSP, but the proposed CPSP method beats both of them. Take the VGGSound-AVEL100k for an example, compared to the PSP, the average performance of CPSP exceeds it by 1.9\% (from 57.8\% to 59.7\%) and 1.1\% (from 47.1\% to 48.2\%) for the fully and weakly supervised settings, respectively.
\textbf{Third}, the proposed CPSP keeps high performance but with relatively low accuracy variance (Accuracy@std) that show its robustness.
Notably, to keep consist with public literature in the main paper, we report the best performance on AVE dataset and the same to VGGSound-AVEL100k dataset.

\begin{table*} [h]
\renewcommand\arraystretch{1.2}
\caption{Comparison of multiple runs on the AVE and VGGSound-AVEL100k datasets in both fully and weakly supervised settings. Accuracy@avg and Accuracy@std denote the averaged and the standard deviation of the accuracy values, respectively.}
\centering
\begin{tabular}{c|p{1.cm}<{\centering}|p{1.5cm}<{\centering}|p{1.cm}<{\centering}p{1.cm}<{\centering}p{1.cm}<{\centering}p{1.cm}<{\centering}p{1.2cm}<{\centering}|p{2.2cm}<{\centering}|p{2.cm}<{\centering}}
\toprule[1pt]
 Dataset & Setting                  & Method & Run-1 & Run-2 & Run-3 & Run-4 & Run-5 &  Accuracy@avg $\uparrow$ &  Accuracy@std $\downarrow$ \\ \hline
\multirow{12}{*}{\rotatebox{90}{AVE~\cite{tian2018audio}}} & \multirow{6}{*}{fully}   & AVEL~\cite{tian2018audio}   & 70.5   & 69.9   & 70.5   & 69.0   & 71.3   & 70.2 & 0.8     \\ 
& & AVSDN~\cite{lin2019dual}* & 70.8   & 72.3   & 72.0   & 71.2   & 69.7   & 71.2  & 0.9   \\ 
& & CMAN~\cite{xuan2020cross}* & 70.7   & 72.3   & 71.7   & 71.2   & 71.3   & 71.4  & 0.5   \\ 
& & CMRA~\cite{xu2020MM}   & 74.8   & 74.1   & 74.7   & 74.6   & 74.3   & 74.5  & 0.3  \\ 
& & PSP~\cite{zhou2021positive}   & {77.8}   & 77.4   & 76.3   & 77.3   & 76.8   & 77.1  & 0.5     \\ 
& & \textbf{CPSP}       & \boxed{\textbf{78.6}}   & \textbf{77.9}   & \textbf{78.0}   & \textbf{78.3}   & \textbf{78.3}   & \textbf{78.2} & \textbf{0.2}  \\ \cline{2-10}
& \multirow{6}{*}{weakly} & AVEL~\cite{tian2018audio}  & 66.1   & 66.2   & 66.0   & 66.3   & 65.0   & 65.9 & 0.5    \\ 
& & AVSDN~\cite{lin2019dual}*  & 65.3   & 62.6   & 61.8   & 62.0   & 58.9   & 62.1 & 2.0    \\ 
& & CMAN~\cite{xuan2020cross}*  & 68.6   & 68.8   & 68.5   & 68.6   & 68.8   & 68.7 & \textbf{0.1}    \\ \
& & CMRA~\cite{xu2020MM}   & 71.2   & 71.3   & 70.7   & 71.6   & 71.8   & 71.3  & 0.4     \\ 
& & PSP~\cite{zhou2021positive}   & {73.5}   & 72.2   & 73.3   & 72.8   & 72.5   & 72.9  & 0.5     \\ 
& & \textbf{CPSP}    & \boxed{\textbf{74.2}}   & \textbf{73.6}   & \textbf{74.0}   & \textbf{73.8}   & \textbf{73.9}   & \textbf{73.9} & 0.2 \\\hline \hline
\multirow{8}{*}{\makecell[l]{\rotatebox{90}{\scriptsize{VGGSound-AVEL100k}}}} & \multirow{4}{*}{fully}   & AVEL~\cite{tian2018audio}   & 55.7   & 55.1   & 54.8                     & 55.6                     & 55.6   & 55.4 & 0.3    \\  
  &    & CMRA~\cite{xu2020MM}* & 57.1   & 56.6   & 56.3                     & 56.7                    & 56.9   & 56.7  & 0.3   \\ 
 &     & PSP~\cite{zhou2021positive}    & {58.3}   & 57.1   & 57.6 & 57.9                     & 58.0   & 57.8 & 0.4    \\ 
  &    & \textbf{CPSP}   & \boxed{\textbf{59.9}}   & \textbf{59.7}   & \textbf{59.8}                    & \textbf{59.5} & \textbf{59.5}   & \textbf{59.7}  & \textbf{0.2}   \\
  \cline{2-10}
& \multirow{4}{*}{weakly} & AVEL~\cite{tian2018audio}  & 45.5  & 46.0   & 45.6                     & 45.7                     & 46.2   & 45.8 & 0.3    \\ 
& & CMRA~\cite{xu2020MM}* & 46.8   & 46.6   & 46.3                    & 46.0                     & 46.8  & 46.5 & 0.3  \\ 
& & PSP~\cite{zhou2021positive} & {47.4}   & 46.7   & 47.4                    & 46.5                    & 47.4   & 47.1 & 0.4     \\ 
& & \textbf{CPSP}   & {\textbf{48.3}}   & \textbf{48.2}   & \textbf{47.8}                     & \textbf{48.3}                    & \boxed{\textbf{48.4}}   & \textbf{48.2} & \textbf{0.2}
\\ \bottomrule[1pt]
\end{tabular}
\begin{tablenotes}
\label{table:avg_result_on_AVE_VGGSound}
\footnotesize
\item[*] * indicates that method is reproduced by us. $\square $ marks the best performance in the multi-runs.
\end{tablenotes}
\end{table*}

\subsection{More discussion on the comparison to self-supervised methods}\label{supp_sspsp}
\textbf{Comparison to the SSPSP.} As introduced in the main paper, SSPSP directly takes the synchronized audio-visual segment pair as a positive sample and otherwise is negative. 
For the self-supervised SSPSP, we observe the new results from three aspect as follows.

\emph{1) SSPSP uses a huge order of magnitude more $<$postive, negative$>$ samples} than CPSP for contrastive learning.
Given a batch of video data, we denote the number of videos in this batch as $N^b$. In SSPSP, for each audio segment, all the visual segments from the batch except the synchronized segment (\emph{including all the intra- and inter- video segments in a batch}) are treated as negative samples. The number of negative samples is $N^b T \times (N^b T - 1)$, where $T$ denotes each video contains $T$ video segments. In contrast, the CPSP merely adopts \emph{background segments in a video} as negative samples.
Therefore, the negatives used in CPSP is just $N^b\times (T-n^{bg}) \times n^{bg}$, where $n^{bg}$ is the number of background segments per video. There is ($N^b T - 1) \gg n^{bg}$. 
Obviously, the amount of negative samples used in the SSPSP is a huge order of magnitude more than that in CPSP.
For example, by statistics, when extending to the whole dataset, the number of negative samples at segment-level used during training of SSPSP and CPSP is approximately $1537 : 1$ on the AVE dataset.
As a result, SSPSP achieves promising performance that attributes the success to data-driven with numerous negative samples.
However, this cannot completely dispel the doubt that SSPSP sounds not solidly and technically for the AVE Localization task. The doubt is that SSPSP will inevitably bring false negatives of audio-visual pair depicting the same event but at different timestamp. Theoretically, these audio-visual pairs are audio-visual correspondence but will be wrongly considered as the negatives.

2) \emph{SSPSP benefits from subset $\mathcal{D}_{\text{bg}}$ 
but distorts from $\mathcal{D}_{\text{ae}}$ in each dataset}. 
We examine the methods on the data distribution.
As shown in Table~\ref{table:SSPSP_on_subset}, SSPSP performs much better on the video type $\mathcal{D}_{\text{bg}}$ 
than $\mathcal{D}_{\text{ae}}$. $\mathcal{D}_{\text{bg}}$ denotes the video that contains both AVE and background segments (as shown in Fig.~\ref{fig:subset_bg});  $\mathcal{D}_{\text{ae}}$ represents another type of video that contains pure AVE segments belonging to a certain event category (as shown in Fig.~\ref{fig:subset_ae}). 
For example, on the large-scale VGGSound-AVEL100k dataset, the SSPSP has terrible performance drop compared with the proposed CPSP from 58.8\% $\rightarrow$ 50.2\% $\downarrow$ with $\mathcal{D}_{\text{ae}}$. 
It reflects the weakness of such self-supervised learning in theory - taking the unsynchronized audio-visual segment pairs but still semantic-corresponding (positive samples in fact) as the negatives introduces extra noises for the AVE localization task.
The performance of SSPSP obviously drops in the case of fine-tuning with $\mathcal{D}_{\text{ae}}$ compared to $\mathcal{D}_{\text{bg}}$. Unlike this, the CPSP performs stably and is almost not affected under all the conditions.

3) \emph{SSPSP is sensitive to the training batch size.}
As well known, the performance of such self-supervised method is influenced by the number of the negative samples during training. 
We test the influence of batch size here. As shown in Table~\ref{table:SSPSP_BatchSize}, the performance of the SSPSP obviously decreases under the small batch size (58.8\% with batch size 128 $\rightarrow$ 55.8\% with batch size 32). Unlike this, the CPSP keeps highly stable.

To summarize, such self-supervised learning can bring some improvements for AVEL by learning from enormous audio-visual pairs but it is much more sensitive to the data distribution and the training batch size. Facing these factors, the CPSP is still superior and robust.

\begin{table}[h]
  \caption{Evaluation on each subset for the CPSP and the SSPSP in the fully supervised setting, measured by accuracy(\%) on the AVE and VGGSound-AVEL100k datasets.}
    \begin{center}
\begin{threeparttable}
  \begin{tabular}{lp{.6cm}<{\centering}p{.6cm}<{\centering}p{1.4cm}<{\centering}p{.6cm}<{\centering}p{.6cm}<{\centering}p{1.4cm}<{\centering}}
  \toprule[0.8pt]
  \multirow{3}{*}{Method}  & \multicolumn{3}{c}{AVE}  & \multicolumn{3}{c}{VGGSound-AVEL100k} \\ 
  &\multicolumn{3}{c}{($\#\mathcal{D}_{\text{ae}}:\#\mathcal{D}_{\text{bg}} \approx$ 2:1)} & \multicolumn{3}{c}{($\#\mathcal{D}_{\text{ae}}:\#\mathcal{D}_{\text{bg}} \approx$ 3:2)}\\ \cmidrule(r){2-4}\cmidrule(r){5-7}
  & $\mathcal{D}_{\text{ae}}$ & $\mathcal{D}_{\text{bg}}$ &  $\mathcal{D}_{\text{ae}}$ \& $\mathcal{D}_{\text{bg}}$ & $\mathcal{D}_{\text{ae}}$ & $\mathcal{D}_{\text{bg}}$  & $\mathcal{D}_{\text{ae}}$ \& $\mathcal{D}_{\text{bg}}$  \\ \midrule
  PSP   & -  & -  & 77.8  & - & - & 58.3  \\
  SSPSP   & 77.4  & 77.9  & 78.2  & \ul{50.2}  & 57.6 & 58.8  \\
  CPSP    & \textbf{78.2}  & \textbf{78.3}  & \textbf{78.6}  & \textbf{58.6}  & \textbf{59.2} & \textbf{59.9} \\ \bottomrule[0.8pt]
  \end{tabular}
  \footnotesize
  $^*\#\mathcal{D}_{\text{bg}}$ denotes the size of subset $\mathcal{D}_{\text{bg}}$, and the same is $\#\mathcal{D}_{\text{ae}}$.
  \end{threeparttable}
     \end{center}
  \label{table:SSPSP_on_subset}
 \vspace{-3mm}
\end{table}

\begin{table}[h]
\setlength{\tabcolsep}{4mm}
   \caption{Performances of the CPSP and SSPSP with different training batch-sizes on the VGGSound-AVEL100k dataset in the fully supervised setting,
   measure by the accuracy(\%).}
   \begin{center}
   \begin{tabular}{{lp{1.5cm}<{\centering}p{1.5cm}<{\centering}p{1.5cm}<{\centering}p{1.5cm}<{\centering}}}
   \toprule[0.8pt]
    \multirow{2}{*}{Method} &\multicolumn{3}{c}{Batch size } \\ \cline{2-4}
    & 128  & 64 & 32 \\ \hline
   SSPSP   & 58.8 & 57.7 & 55.8  \\ 
   CPSP & \textbf{59.9} & \textbf{59.7} & \textbf{59.5}\\ \bottomrule[0.8pt]
   \end{tabular}
   \end{center}   \label{table:SSPSP_BatchSize}
   \vspace{-3mm}
\end{table}

\noindent \textbf{Comparison to the self-supervised method Global-Local~\cite{ma2021contrastive}.}
Global-Local~\cite{ma2021contrastive}
is built upon the large-scale pre-training and achieves remarkable performance on the downstream audio-visual event localization task. It performs both spatial and temporal audio-visual contrastive learning in a self-supervised manner. 
To compare with Global-Local~\cite{ma2021contrastive}, we adopt the same visual backbone, \ie, the advanced 3D-ResNet, to extract the visual features.
The results are shown in Table~\ref{table:CPSP_with_different_visual_features},  our method is comparable to the Global-Local~\cite{ma2021contrastive}. 
For the fully supervised setting, the proposed CPSP with segment-level and video-level positive sample activation (PSA$_S$ and PSA$_V$) achieves better performance than Global-Local~\cite{ma2021contrastive} (82.3\% vs. 82.1\%).
As for the weakly supervised setting, the CPSP equipped with only video-level PSA$_V$ is comparable to Global-Local~\cite{ma2021contrastive} (78.7\% vs. 79.8\%).

But importantly, our method achieves such comparable performances at a much lower cost. 
1) The Global-Local~\cite{ma2021contrastive} is 
pretrained on an extra large-scale K-AV-240K dataset that contains 240k audio-visual videos, while our method does not need any extra data. 
2) To achieve better performances, the Global-Local~\cite{ma2021contrastive} samples video frames at 10 FPS on AVE~\cite{tian2018audio} dataset for data augmentation. In our work, in order to keep the consistent experiment setup, we keep 1 FPS in our method as the same as existing literature~\cite{tian2018audio,lin2019dual,xu2020MM,xuan2020cross,wu2019dual,ramaswamy2020see,ramaswamy2020makes,yu2021mpn}. 
3) The authors of Global-Local~\cite{ma2021contrastive} suggest using 16 Tesla P100 GPU to handle the large-scale dataset of 240k videos while our model can be trained very lightly with just one GTX 1080 GPU without extra data. Consequentially, it is obvious that our method spends less training time and fewer GPUs since it uses much fewer data.

To summarize, even without those techniques used in the Global-Local~\cite{ma2021contrastive} (\emph{e.g.}, large-scale pre-training, dense video frames sampling, high-performance GPU device), our model still performs competitively.

\begin{table}[H]
   \caption{Comparison with the self-supervised Global-Local built upon large-scale pre-training~\cite{ma2021contrastive} for audio-visual event localization.}
   \begin{center}
\begin{threeparttable}
   \begin{tabular}{lp{2.2cm}<{\centering}p{1.8cm}<{\centering}p{2.0cm}<{\centering}p{1.6cm}<{\centering}p{1.2cm}<{\centering}p{1.2cm}<{\centering}}
   \toprule[1pt]
   \multirow{3}{*}{Method} & \multicolumn{2}{c}{Dataset} & \multirow{3}{*}{\#GPU} & \multirow{3}{*}{\makecell[c]{Visual\\Encoder}}   & \multicolumn{2}{c}{Setting} \\ \cmidrule(r){2-3} \cmidrule(r){6-7}
   & extra data & train data & & & fully & weakly \\ \midrule
  \multirow{2}{*}{CPSP}   & \multirow{2}{*}{none} & \multirow{2}{*}{AVE~\cite{tian2018audio}]} & \multirow{2}{*}{{1 \small{GTX 1080 }}} & VGG-19 &  78.6 & 74.2 \\ \cmidrule(r){5-7}
   &  &  &    & {3D-ResNet} & \textbf{82.3} & 78.7 \\   \midrule
 Global-Local~\cite{ma2021contrastive} & K-AV-240K~\cite{ma2021contrastive}  & \makecell[c]{AVE~\cite{tian2018audio}} & {16 \small{Tesla P100}} & {3D-ResNet} & 82.1 &  \textbf{79.8} \\
 \bottomrule[1pt]
   \end{tabular}
   \end{threeparttable}
   \end{center}
   \label{table:CPSP_with_different_visual_features}
 \vspace{-1mm}
\end{table}

\section{Experiments for Audio-Visual Video Parsing.}\label{supp_avvp}
\textbf{Ablation study on the threshold $\mu$.}
As introduced in the main paper, the threshold $\mu$ is used to select the positive samples for the AVVP task. 
We perform an ablation study on $\mu$ and the results are shown in Table~\ref{table:CPSP_ratio_r}.
The best performance of CPSP is achieved when $\mu$ is set to 0.6. When $\mu$ is set to 1.0, it means only videos have the exactly same label set of multiple instance categories will be selected as positives. This condition is kind of strict especially in a batch video data during training.
We argue that $\mu=0.6$ (60\% category coverage) is a suitable parameter to make the model flexible in selecting positive samples during training and we choose this setup in the experiments for AVVP.
We have also released the pretrained model at \href{https://drive.google.com/drive/folders/1cqlVeAKx1NFKnk0ynXvyyQHnnKqyecNm?usp=sharing}{\emph{https://drive.google.com/drive/folders/1cqlVeAKx1NFKnk0ynXvyyQHnnKqyecNm?usp=sharing}}.

\begin{table*}[h]
\caption{Ablation study on the threshold $\mu$ used for $\mathcal{L}_{\text{vpsa}}$ to construct positive and negative samples for AVVP task. Experiments are conducted on the LLP dataset.}
   \begin{center}
\begin{threeparttable}
   \begin{tabular}{lp{1.2cm}<{\centering}p{1.2cm}<{\centering}p{1.2cm}<{\centering}p{1.4cm}<{\centering}p{1.5cm}<{\centering}p{1.2cm}<{\centering}p{1.2cm}<{\centering}p{1.2cm}<{\centering}p{1.4cm}<{\centering}p{1.5cm}<{\centering}}
   \toprule[0.8pt]
   \multirow{3}{*}{$\mu$} & \multicolumn{5}{c}{Segment-level} & \multicolumn{5}{c}{Event-level}\\ \cmidrule(r){2-6} \cmidrule(r){7-11}
   & A & V & AV & Type@AV & Event@AV & A & V & AV & Type@AV & Event@AV\\ \midrule
  0.5 & 58.2 & 56.3 & 51.4 & 55.3 & 55.5 & 51.0 & 52.4 & 45.3 & 49.6 & 49.3 \\
  \textbf{0.6} & \textbf{58.5} & \textbf{57.8} & \textbf{52.6} & \textbf{56.3} & \textbf{55.8} & \textbf{51.6} & \textbf{54.0} & \textbf{46.5} & \textbf{50.7} & \textbf{49.9} \\
  1.0 & 56.2 & 56.8 & 50.2 & 54.4 & 54.8 & 48.9 & 52.3 & 44.0 & 48.4 & 48.0 \\
 \bottomrule[0.8pt]
   \end{tabular}
   \end{threeparttable}
   \end{center}
   \label{table:CPSP_ratio_r}
 \vspace{-3mm}
\end{table*}

\end{document}